\pdfoutput=1

\documentclass[11pt]{article}
\usepackage{newtxtext}
\usepackage{ltablex} 

\usepackage[final]{acl_local}
\usepackage{siunitx}
\usepackage{threeparttable}
\usepackage{times}
\usepackage{latexsym}

\usepackage{kotex}
\usepackage[T1]{fontenc}
\usepackage{lmodern}

\usepackage[utf8]{inputenc}

\usepackage{inconsolata}

\usepackage{graphicx}
\usepackage{listings}
\usepackage[most]{tcolorbox}
\tcbuselibrary{listings}
\usepackage{caption}
\usepackage{graphicx}
\usepackage{multicol}
\usepackage{amsmath}
\usepackage{booktabs}
\usepackage{multirow}
\usepackage{tabularx}
\usepackage{makecell}
\usepackage{subcaption}
\usepackage{arydshln}

\usepackage{algorithm}
\usepackage{algorithmic}
\usepackage{xcolor}
\usepackage[table]{xcolor}
\usepackage{longtable}

\definecolor{PastelPink}{HTML}{E31E1E}

\definecolor{PastelPurple}{HTML}{6621D4}

\definecolor{PastelGreen}{HTML}{63F1C3}

\definecolor{PastelBlue}{HTML}{25C9ED}
\definecolor{PastelYellow}{HTML}{F3F364}

\title{\textsc{LegalSearchLM}: Rethinking Legal Case Retrieval as Legal Elements Generation
}

\author{
  Chaeeun Kim$^{1}$\thanks{Correspondence to: \href{mailto:email@domain}{chaechaek1214@gmail.com}} \quad Jinu Lee$^{2}$\quad Wonseok Hwang$^{1,3}$ \\ \\ 
$^{1}$LBOX\hspace{0.5cm}$^{2}$University of Illinois Urbana-Champaign\hspace{0.5cm}$^{3}$University of Seoul\\
}

\begin{document}
\maketitle
\begin{abstract}
Legal Case Retrieval (LCR), which retrieves relevant cases from a query case, is a fundamental task for legal professionals in research and decision-making. However, existing studies on LCR face two major limitations. First, they are evaluated on relatively small-scale retrieval corpora (e.g., 100-55K cases) and use a narrow range of criminal query types, which cannot sufficiently reflect the complexity of real-world legal retrieval scenarios. Second, their reliance on embedding-based or lexical matching methods often results in limited representations and legally irrelevant matches.
To address these issues, we present: (1) \textbf{\textsc{LEGAR BENCH}}, the first large-scale Korean LCR benchmark, covering 411 diverse crime types in queries over 1.2M candidate cases; and (2) \textbf{\textsc{LegalSearchLM}}, a retrieval model that performs legal element reasoning over the query case and directly generates content containing those elements, grounded in the target cases through constrained decoding. Experimental results show that \textsc{LegalSearchLM} outperforms baselines by 6 -- 20\% on \textsc{LEGAR BENCH}, achieving state-of-the-art performance. It also demonstrates strong generalization to out-of-domain cases, outperforming naive generative models trained on in-domain data by 15\%.
\end{abstract}

\begingroup
\setlength{\tabcolsep}{6pt}
\renewcommand{\arraystretch}{0.85}

\begin{table*}[t!]
\centering
\fontsize{9}{14}\selectfont
\resizebox{0.85\textwidth}{!}{
    \begin{tabular}{cccccccc}
        \toprule
              & Language & Crime types of query & Query case & Retrieval pool & Target case per query \\
    \midrule
    
    $\text{COLIEE2024}$ & $\text{English}$ & - & 400 & 1,734 (per query) & - \\
    $\text{LeCaRD}$ & $\text{Chinese}$ & - & 107 & 100 (per query) & 10.33 \\
    $\text{LeCaRDv2}$ & $\text{Chinese}$ & - & 800 & 55,192 (per query) & 20.89 \\
    \midrule
    \textbf{LEGAR BENCH$_{\textit{Standard}}$} & $\text{Korean}$ & 411 & 411*N  & 1,226,814 & 200 \\
    \textbf{LEGAR BENCH$_{\textit{Stricter}}$} & $\text{Korean}$ & 160 & \textasciitilde 15,777  & 169,230 & 14.69 \\
    \bottomrule
    \end{tabular}
    }
\caption{Scale comparison of LCR benchmarks. \textsc{LEGAR BENCH} expands the retrieval pool and query coverage. $N$ indicates query set expandability via similar case groups.}
\label{tabs:comparison_dataset}
\end{table*}

\endgroup 
\section{Introduction}
Legal AI has increasingly gained attention from legal professionals to raise the productivity of their work. Among various legal applications, \textit{Legal Case Retrieval} (LCR)~\cite{feng-etal-2024-legal, deng2024learninginterpretablelegalcase, su2024caseformerpretraininglegalcase,
deng-etal-2024-element,
li2023sailerstructureawarepretrainedlanguage, xiao2021lawformerpretrainedlanguagemodel,li2023lecardv2largescalechineselegal, li2023thuircoliee2023incorporatingstructural, 10.1145/3580305.3599273}, which identifies relevant precedents for a given case, plays a particularly crucial role in maintaining judicial fairness and supporting the decision-making process of legal experts. 

However, previous work on the legal case retrieval task has clear limitations.
(1) Most evaluations have been conducted in small-scale settings, where a predefined set of candidate documents is provided for each query and the queries do not sufficiently reflect the diverse criminal case types~\cite{10.1145/3404835.3463250, li2023lecardv2largescalechineselegal, li2023thuircoliee2023incorporatingstructural, xiao2018cail2018largescalelegaldataset, Li_2023}.
(2) The tested retrieval models are limited to embedding-based similarity and lexical matching approaches, with the former struggling to capture the rich semantics of legal literature as it compresses complex documents into fixed-size vectors, and the latter often leading to unfocused matches due to a lack of semantic understanding~\cite{wang2023neuralcorpusindexerdocument, magesh2024hallucinationfreeassessingreliabilityleading, kim-etal-2024-exploring-practicality}.


To address the first, we present \textsc{LEGAR BENCH} (\textbf{Lega}l Case \textbf{R}etrieval \textbf{Bench}mark), the first large-scale Korean LCR benchmark, comprising two dataset versions tailored to different evaluation needs:
(1) LEGAR BENCH$_{\textit{Standard}}$ is designed for comprehensive assessment across a broad range of crime categories. The queries cover 411 distinct crime types and are evaluated over a retrieval pool of 1.2M cases. To achieve this, we systematically construct the benchmark using crime types based on statutory provisions, even further detailed than official charge titles used in courts (e.g., defamation by fact disclosure – Article 307(1), or false allegation – Article 307(2), both sharing the charge title \textit{defamation}) (Section~\ref{sec:standard_bench}).
(2) LEGAR BENCH$_{\textit{Stricter}}$ evaluates stricter relevance criteria than LEGAR BENCH$_\textit{Standard}$. It considers more factual details and legal issues within the same crime type that can affect the final judgment or sentence, which is crucial for legal practitioners. To enable this, we annotate the pool of 170K cases across 160 crime types using 102 crime-specific legal factors and 443 corresponding options (Section~\ref{subsec:legar_sticter}).



To address the second, we shift our focus on the LCR task from sequence-to-sequence matching to generating the key legal elements that determine relevance by proposing \textsc{LegalSearchLM}. It directly generates important legal elements via constrained decoding and returns the documents that contain them~\citep{li2024matchinggenerationsurveygenerative, kim-etal-2024-exploring-practicality, li2023learningrankgenerativeretrieval, li-etal-2023-multiview, bevilacqua2022autoregressivesearchenginesgenerating}, mitigating the shortcomings of traditional retrievers like BM25 and dense embeddings. Given the specificity of the legal domain, we adopt diverse strategies including first token-aware generation and self-supervised fine-tuning (SSFT) (Section~\ref{sec:legal_search_lm}). 

We evaluate \textsc{LegalSearchLM} on \textsc{LEGAR BENCH}, comparing it against strong baselines including lexical matching and embedding-based methods from both general-purpose and legal-domain models. Our experimental results show that \textsc{LegalSearchLM} outperforms the best baseline by 6\% in precision. Our training strategy leads to significantly better generalization, achieving a 15\% improvement over generative retrieval models trained with naive identifiers on in-domain data (Section \ref{sec:analysis_discussion}).

In summary, our contributions are as follows.
\begin{itemize}
    \item We introduce the first Korean LCR benchmark, \textsc{LEGAR BENCH}, which has the largest and most diverse criminal cases.
    \item We present \textsc{LegalSearchLM} that generates legal elements that should be included in target case via first-token-aware decoding.
    \item Our \textsc{LegalSearchLM} achieves state-of-the-art performance on LEGAR BENCH, demonstrating remarkable \textit{generalization ability} on unseen crime types.
\end{itemize}

\section{\textsc{LEGAR BENCH}}
\label{sec:legalbench}
\begingroup
\setlength{\tabcolsep}{6pt}
\renewcommand{\arraystretch}{0.8}

\begin{table}[t!]
    \centering
    \tiny

    \fontsize{8}{11}\selectfont
    \resizebox{0.48\textwidth}{!}{
        \begin{tabular}{cccc}
            \toprule
            \multirow{2}{*}{Categories} & \multicolumn{2}{c}{\#Crime types of queries} \\
            \cmidrule(lr){2-3}
            & Standard & Stricter \\
             \midrule
            Traffic offenses & 13 & 9 \\
            Fraud & 21 & 8 \\
            Injury or Violence & 31 & 19 \\
            Sexual crime & 132 & 111 \\
            Defamation or Insult & 8 & 6 \\
            Finance or Insurance & 5 & 1 \\
            Drug & 5 & 4 \\
            Murder & 2 & 2 \\
            Theft or Robbery & 38 & - \\
            Obstruction of Business & 13 & - \\
            Destruction & 5 & - \\
            Threat & 11 & - \\
            Criminal trespass & 15 & - \\
            Embezzlement or Breach of trust & 15 & - \\
            Gambling & 7 & - \\
            Negligent homicide and injury & 6 & - \\
            Obstruction of right & 5 & - \\
            Child abuse or School violence & 10 & - \\
            Medical or Food drug & 11 & - \\
            Corporation & 3 & - \\
            Bribery & 3 & - \\
            Car & 2 & - \\
            Labor or Employment & 11 & - \\
            Industrial or Serious accidents & 4 & - \\
            Military duty or law & 2 & - \\
            Consumer or Fair trade & 1 & - \\        
            Arrest or Detention & 1 & - \\     
            Intellectual property & 3 & - \\
            IT or Privacy & 2 & - \\
            Misdemeanor & 1 & - \\
            Sexual norms & 1 & - \\
            Tax, Administ, Const law & 14 & - \\
            Other criminal offenses & 10 & - \\
            \midrule
            \textbf{Total} & \textbf{411} & \textbf{160} \\
            \bottomrule
        \end{tabular}
    }
    \caption{Statistics of Crime Typology and Corresponding Criminal Types. LEGAR BENCH$_\textit{Standard}$ includes 411 query types across 33 crime categories, while LEGAR BENCH$_\textit{Stricter}$ covers 160 query types across 8 categories.} 
    \label{tables:standard_stricter_stats}
\end{table}

\endgroup

In this section, we describe relevance criteria and the construction process of \textsc{LEGAR BENCH}.

\textsc{LEGAR BENCH} features the most comprehensive set of query and target cases (See Table \ref{tabs:comparison_dataset}), with relevance criteria rigorously defined by the lawyers involved in the construction process. It offers two dataset versions based on different evaluation needs, LEGAR BENCH$_{\textit{Standard}}$ and LEGAR BENCH$_{\textit{Stricter}}$.

\begin{figure*}[t]
{
\includegraphics[width=1\linewidth]
{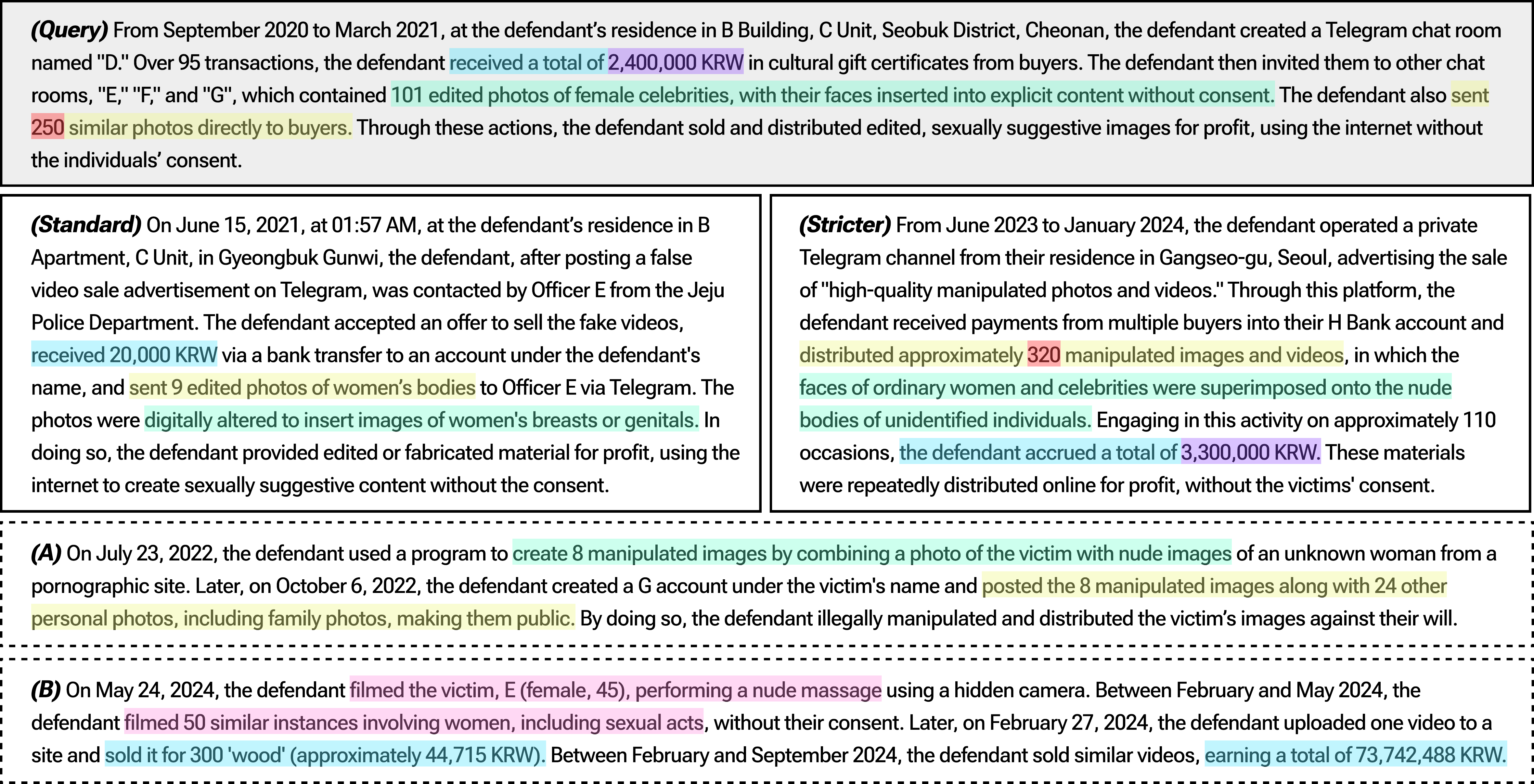}
}
\caption{Examples of Relevance Cases. \textbf{(Query)} is a query case on distributing false images/videos for profit. The \textcolor{PastelBlue}{Blue Highlight} indicates profit, the \textcolor{PastelYellow}{Yellow Highlight} represents the creation of false images/videos, and the \textcolor{PastelGreen}{Green Highlight} denotes distribution—the three key legal elements of the crime. Both \textbf{(Standard)} and \textbf{(Stricter)} satisfy the three elements, and \textbf{(Stricter)} additionally meets the requirements concerning the scale of distributed images/videos \textcolor{PastelPink}{(Red Highlight)} and the total financial gains obtained \textcolor{PastelPurple}{(Purple Highlight)}. \textbf{(A)} and \textbf{(B)} are not target cases, as \textbf{(A)} distributed a false image without intending to obtain financial gains, and \textbf{(B)} committed the offense for financial gain through the unlawful filming of real footage, not the creation of false images.}
\label{fig:examples_standard_stricter}
\end{figure*}

\begin{figure}[t!]
{
\centering
    \begin{minipage}[b]{0.49\textwidth}
    \centering
    \includegraphics[width=\textwidth]
    {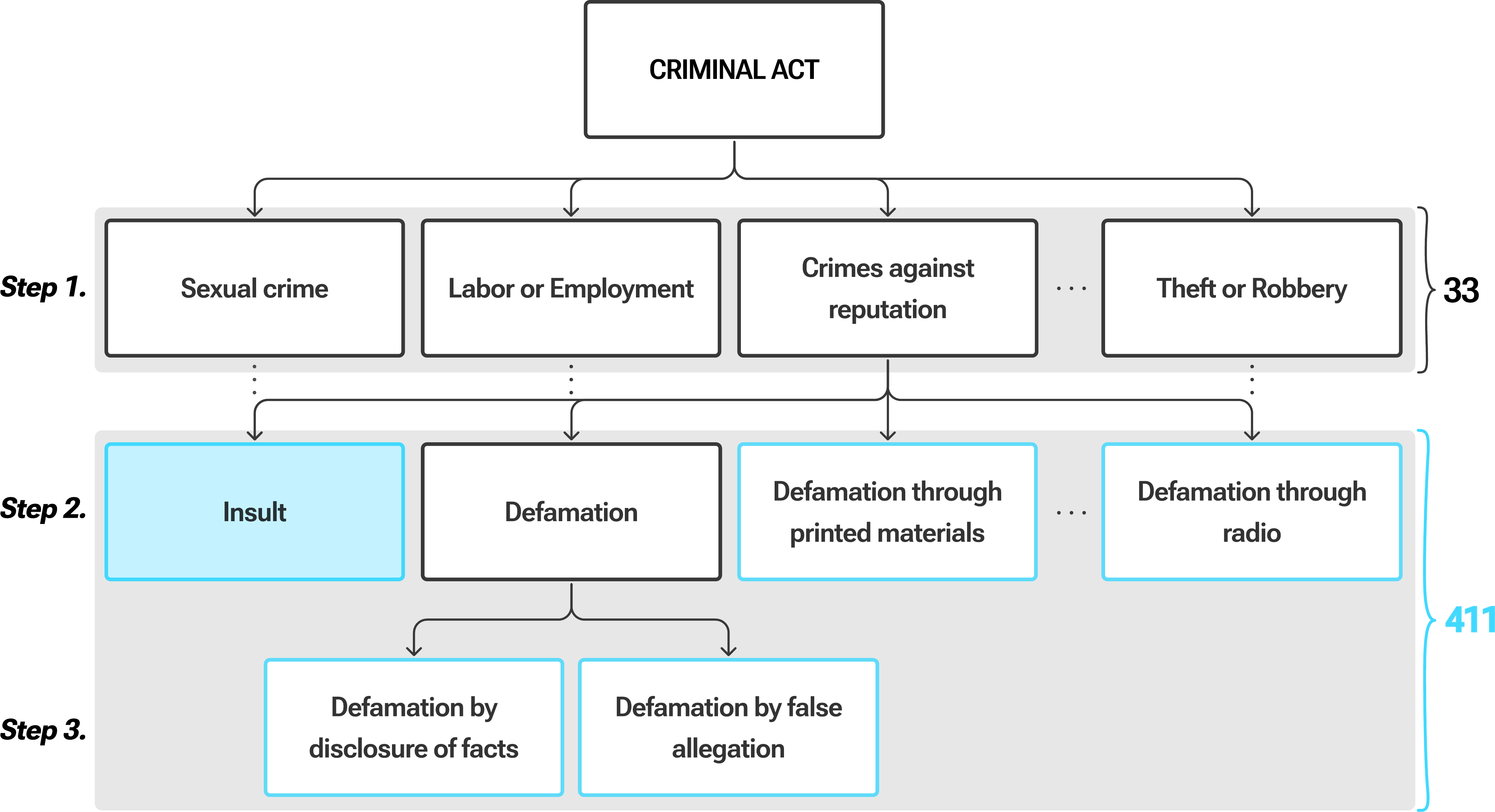}
    \caption{Examples of the construction process for each step in LEGAR BENCH$_{\textit{Standard}}$. \textbf{Step 1} defines major crime categories based on Korean Criminal Act. \textbf{Step 2} refines these categories using charge titles, and \textbf{Step 3} further specifies them based on statutory provisions.}
    \label{fig:d1_category}
    \end{minipage}
}
\end{figure}

\subsection{\texorpdfstring{LEGAR BENCH$_{\textit{Standard}}$}{LEGAR BENCH (Standard)}}

\label{sec:standard_bench}
LEGAR BENCH$_{\textit{Standard}}$ is designed to provide a comprehensive assessment of retrieving diverse crime categories, and the enhancement of overall system performance.
\label{subsec:legar_standard}
\subsubsection{Definition of Standard Relevance}
We define standard relevance based on the charge title and statutory provision, where the former refers to the formal name of the offense, and the latter indicates the specific statutory article applicable to that charge. For example, as shown in Figure \ref{fig:examples_standard_stricter} on sexual crime, for the query case \textit{\textbf{(Query)}} on distributing false sexual images/videos for profit, the standard target case \textit{\textbf{(Standard)}} satisfies the three statutory elements: 1. creation of false sexual images/videos, 2. intent for profit, and 3. distribution. Cases like \textit{\textbf{(A)}}, which are not for profit, or \textit{\textbf{(B)}}, which concern illegal sexual video filming rather than false sexual image creation, cannot be considered target cases, since they are distinct crimes governed by different laws.

\subsubsection{Data Construction}
To ensure comprehensive coverage of the diverse and complex legal literature, we employ a top-down approach, systematically categorizing crimes based on Korean Criminal Act.

\paragraph{Step 1: Construction of Crime Typology.} We establish a crime typology to categorize various types of crimes in criminal cases. As shown in Step 1 of Figure \ref{fig:d1_category}, we define major categories based on the structure of the Korean Criminal Act, such as sexual crimes, labor or employment offenses, crimes against reputation, and theft or robbery. Appendix \ref{tables:standard_stats} lists a total of 33 crime categories.

\begin{figure}[t!]
{
\centering
    \begin{minipage}[b]{0.49\textwidth}
    \centering
    \includegraphics[width=\textwidth]
    {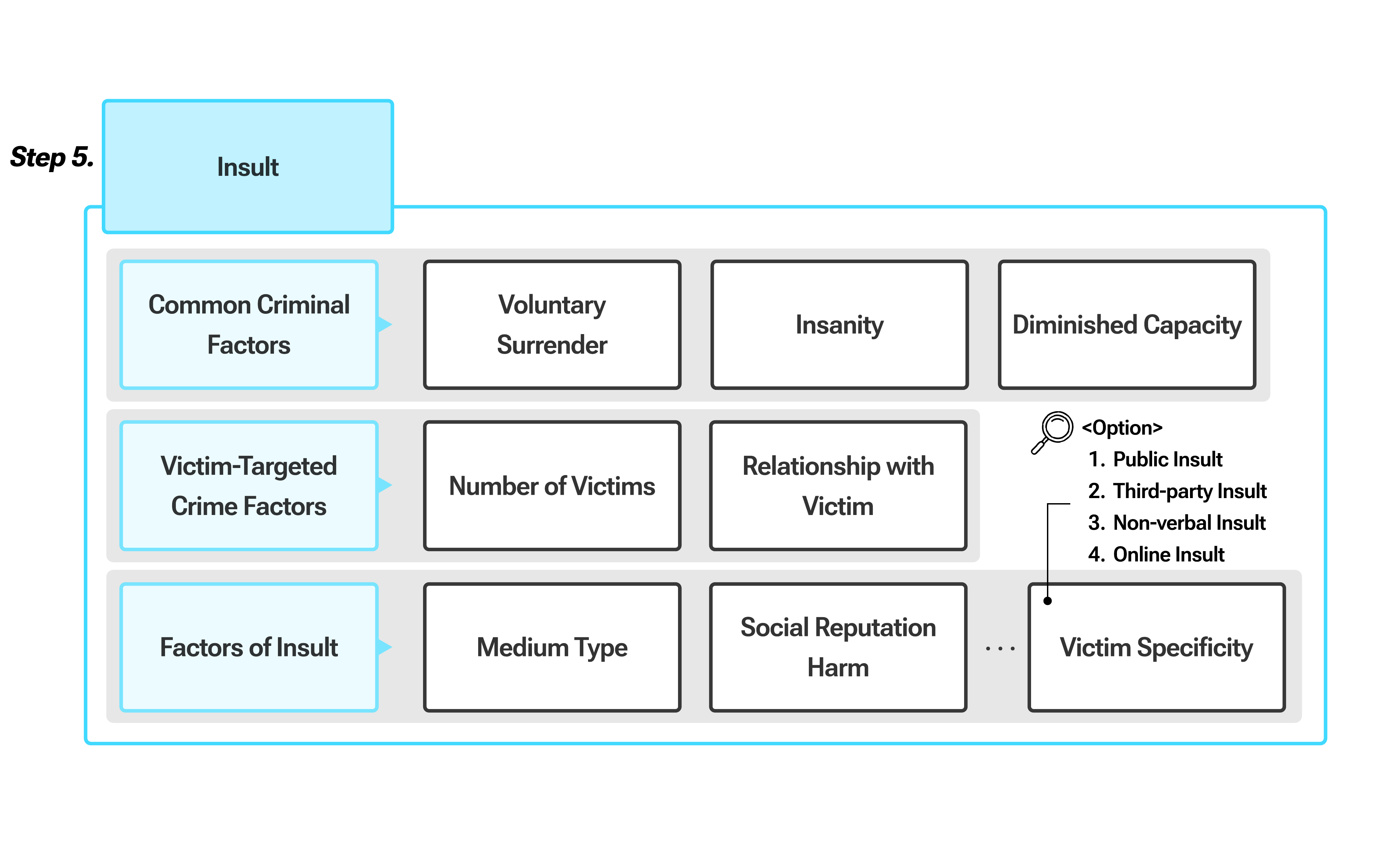}
    \caption{Examples of the construction process of LEGAR BENCH$_{\textit{Stricter}}$. Each crime type (e.g., Insult) includes specifically defined factors (in boxes filled with sky blue) and sub-factors (in boxes outlined in black). Cases are annotated by mapping all sub-factors to corresponding predefined options.}
    \label{fig:stricter_step}
    \end{minipage}
}
\end{figure}

\paragraph{Step 2: Assignment of Charge Titles.}
We construct the set of crime charge titles that can occur within each crime category.
A charge title is the official name used in legal documents, such as indictments or complaints, to describe a specific offense. As shown in Figure \ref{fig:d1_category}, crimes against reputation can be expanded into related charge titles (sub-categories), such as defamation, defamation through printed materials, defamation through radio, insults, etc. While charge titles are determined by statutory provisions that apply, there is no one-to-one correspondence between these two because some charge titles correspond to multiple provisions with subtle differences. This ambiguity is resolved by further refinement in the next step. 

\paragraph{Step 3: Refinement from Statutory Provisions.}  
To better reflect the subtle difference of the body of the crime between provisions grouped within the same charge title, we refine charge titles to the level of individual statutory provisions when such refinement is both possible and meaningful. Figure \ref{fig:d1_category} shows how defamation can be divided according to distinct laws, such as defamation by disclosure of facts and defamation by allegation of false facts. Finally, the standard similar groups are formed by combining the results of Step 2 and Step 3, as shown in the skyblue-bordered box of Figure \ref{fig:d1_category}. \textit{As a result, LEGAR BENCH$_{\textit{Standard}}$ contains 411 similar groups across 33 categories.}

\paragraph{Step 4: Case Mapping.}  
We automatically process 1.2M (1,226,814) criminal cases, mapping them to their respective groups based on the charge title and statutory provisions annotated for each group. This process successfully maps 1M cases (1,052,506), which account for 85.79\% of our total criminal cases, \textit{enabling evaluation on the majority of criminal cases through our LEGAR BENCH$_{\textit{Standard}}$.}

\subsection{\texorpdfstring{LEGAR BENCH$_{\textit{Stricter}}$}{LEGAR BENCH (Stricter)}}

\label{subsec:legar_sticter}
\subsubsection{Definition of Stricter Relevance}
For stricter case similarity, we expand the scope from facts to include claims, reasoning, sentencing factors, and conclusions sections from the case, aiming to provide a more comprehensive view of the process of judges. Stricter relevance further requires \textit{factual details} that do not affect the type of charge, but might affect the final judgment or the sentence. Examples include the severity of the crime, the relationship between the defendant and the victim, situational information, and arguments made by defendants. For instance, while making only a few fake images and selling them for 20 dollars is ruled under the same crime with making hundreds of fake videos with thousands of dollars of profit, the stricter factual relevance between the two cases is low (See the red and purple highlights in Figure \ref{fig:examples_standard_stricter}). Also, if two assault defendants make the same claim of self-defense but only one of them is accepted by the judge, these two cases should also be distinguished. Five legal experts have annotated these important factors that determine the stricter relevance between cases (Appendix \ref{appendix:annotation}).

\subsection{Dataset Construction}
\paragraph{Step 5: Define Detailed Factors.}
We construct LEGAR BENCH$_{\textit{Stricter}}$ starting from 160 similar groups across 8 crime categories in LEGAR BENCH$_{\textit{Standard}}$. First, we define sets of factors to be further considered for each group in the standard set. Figure \ref{fig:stricter_step} shows that the standard group ``Insult'' is associated with Common Criminal Factors, Victim-Targeted Crime Factors, and Factors of Insult.
Next, we identify detailed sub-factors and create options for each of them. Finally, based on the defined factors, sub-factors, and options for each standard group, we annotate the cases belonging to each standard group using GPT-4o. The full list of sub-factors is shown in Table \ref{tab:stricter_list}.  



\paragraph{Step 6: Case Grouping.} 
As a result of Step 5, we obtain (sub-factor, option) pairs for each case across all sub-factors required for each standard group. The following grouping algorithm is then applied to find cases with the highest factual relevance. We created one stricter group for each standard group, resulting in 160 queries (See Table \ref{tabs:comparison_dataset} for details).

\begingroup
\setlength{\tabcolsep}{6pt}
\renewcommand{\arraystretch}{0.8}

\begin{algorithm}
\label{alg:strictergroup}
\caption{Stricter Relevance Group}
\begin{algorithmic}[1]
\STATE \textbf{Input:} case\_data, subfactor-option pair\_list
\STATE \textbf{Output:} grouped\_cases

\FOR{each case in case\_data}
    \STATE key = generate\_key(subfactor-option pair\_list) 
    \STATE group[key].append(case)
\ENDFOR

\IF{any group has 2 or more cases}
    \STATE return the group
\ENDIF

\FOR{r = number of subfactors to 1}
    \FOR{each case in case\_data}
        \STATE key = generate\_key(subfactor-option pair\_list[:r])
        \STATE group[key].append(case)
    \ENDFOR
    \IF{any group has 2 or more cases}
        \STATE return the group
    \ENDIF
\ENDFOR

\RETURN None
\end{algorithmic}
\end{algorithm}
\endgroup

\section{\textsc{LegalSearchLM}}
\label{sec:legal_search_lm}
Previous approaches to LCR rely on either embedding-based or lexical matching methods~\cite{magesh2024hallucinationfreeassessingreliabilityleading}. However, embedding models often lose important details by compressing lengthy legal texts into single vectors, while lexical methods struggle to distinguish legally important information from noise. To better reflect how legal experts assess relevance—focusing on specific legal elements that constitute the crime, rather than subtle details like dates, geographic names, or place names—there is a need for a more sophisticated retrieval approach.

In this work, we introduce \textsc{LegalSearchLM}, which addresses the pitfalls of sequence-to-sequence matching by adopting the generative retrieval paradigm~\cite{bevilacqua2022autoregressivesearchenginesgenerating, li-etal-2023-multiview, li2023learningrankgenerativeretrieval, kim-etal-2024-exploring-practicality}. Given a query case, \textsc{LegalSearchLM} generates relevant \textit{legal elements} as keys for the target documents. By modeling retrieval as language modeling, it effectively mitigates the fixed-dimension embeddings' information loss problem and the lexical match's lack of deep semantic understanding, which are both crucial in LCR.


\begin{figure*}[t]
{
\includegraphics[width=1\linewidth]{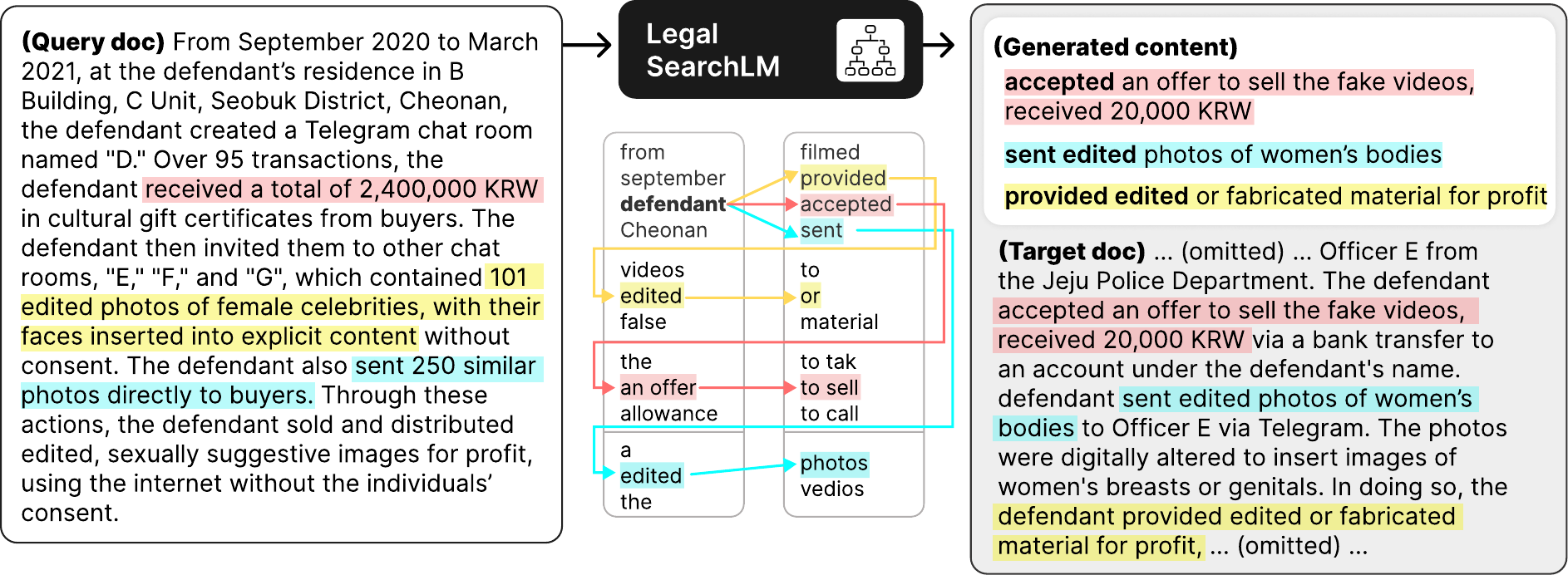}

\caption{Inference process of \textsc{LegalSearchLM}. Given a \textbf{Query doc} as input, \textsc{LegalSearchLM} generates key legal elements expected to appear in the \textbf{Target doc} via \textit{core-first-token-aware} constrained decoding over a prefix-indexed corpus (\textbf{Generated content}). Since the generated content is grounded in the corpus, it can be linked back to its source document, enabling retrieval.}

\label{fig:seachlm_fig}
}
\end{figure*}


\subsection{Training}

\paragraph{Extracting legal elements.} We define legal elements are atomic facts that can influence the final judgment. As proven effective by \citet{min2023factscorefinegrainedatomicevaluation, chen-etal-2024-dense, cai2024textttmixgrenhancingretrievergeneralization}, we extract legally valid elements by prompting an LLM.

However, naively training the model using extracted legal elements is insufficient for successful retrieval. This is especially true at inference time, where the decoding process is constrained by a predefined FM-index that forces the model to generate sequences appearing exactly in the corpus. In this framework, early decoding decisions are highly critical; for instance, generating “dates” or “locations” of the crime as the first token may unintentionally steer the decoding process toward irrelevant paths. This has the unintended effect of filtering based on information unrelated to legal relevance, thereby discarding documents that may contain key legal elements while overemphasizing those that happen to include the specific date or location.

To address this problem, we construct synthetic examples that begin with legally informative tokens from legal elements using few-shot LLMs. Figure~\ref{fig:seachlm_fig} illustrates the generation of first-token-aware legal elements at inference time.



\paragraph{Data collection.} As \textsc{LegalSearchLM} formulates LCR as legal elements generation, it needs finetuning to generate appropriate legal elements given the query case. However, as relevance annotation in the legal domain is costly, training the model on a large set of query-target pairs is not feasible.

As a solution, we construct a training dataset in a self-supervised manner \citep{lewis-etal-2020-bart}. Specifically, we use query case as inputs and legal elements from the \textit{query case} as outputs. 
This approach provides the following three benefits:

\begin{itemize}
    \item \textbf{Less noise and cost-effectiveness:} It reduces noise compared to using an existing retriever's results as gold query-target pairs (e.g., BM25)~\cite{li2023sailerstructureawarepretrainedlanguage}.

    \item \textbf{Balanced training on long-tail crimes:}  It enables better sampling of rare case types than citation-based approaches.

    \item \textbf{Better generalization:} The model learns to reason over legal elements from a query case, aligning with inference-time conditions without relying on memorization.
\end{itemize}

\subsection{Inference}
During inference, \textsc{LegalSearchLM} performs constrained beam decoding over the document index, ensuring that the generated content always exists in a document within the corpus (See Figure~\ref{fig:seachlm_fig}). For constrained decoding, we employ an FM-index based on the Burrows-Wheeler Transform (BWT)~\cite{892127}. It enables efficient exact pattern matching via backward search in time linear to the pattern length. Detailed explanations on the FM-index are in Appendix~\ref{appendix:fmindex}.
The specific generation process is as follows.  

Let \( x_1, x_2, \dots, x_n \) denote a generated token sequence obtained via constrained beam search. At each decoding step \( t \), we maintain a beam of partial sequences \( \{x_{<t}^{(1)}, x_{<t}^{(2)}, \dots, x_{<t}^{(B)}\} \), where each hypothesis is extended based on the previously generated tokens \( x_{<t}^{(i)} \). The candidate set \( \mathcal{C}(x_{<t}^{(i)}) \) represents the valid next tokens for the \( i \)-th hypothesis, constrained by FM-index. This ensures that all generated sequences present within the corpus.



\begin{align}
    x_1^{(i)} &= \arg\max_{x \in V} P(x \mid \text{[BOS]}) 
    \notag \\
    &\text{for } i = 1, \dots, B
    \\
    x_t^{(i)} &\in \text{Top-}k \left\{ P(x \mid x_{<t}^{(i)}) \;\middle|\; x \in \mathcal{C}(x_{<t}^{(i)}) \right\}
    \notag \\
    &\text{for } t \geq 2
\end{align}

In our retrieval process, \textsc{LegalSearchLM} captures key legal elements from the query case that are expected to appear in the target case, and begins generation with informative initial tokens to guide decoding toward relevant parts of the corpus.

\begingroup
\setlength{\tabcolsep}{6pt}
\renewcommand{\arraystretch}{1.5}
\renewcommand{\thefootnote}{\fnsymbol{footnote}}
\begin{table*}[t!]
    \centering
    \renewcommand{\arraystretch}{1.5}
    \sisetup{table-format=2.5, round-mode=places, round-precision=2}
    \begin{threeparttable}
    \resizebox{\textwidth}{!}{
    \begin{tabular}{
      l
      S[table-format=2.2]
      S[table-format=2.2]
      S[table-format=2.2]
      S[table-format=2.2]
      S[table-format=2.2]
      S[table-format=2.2]
      S[table-format=2.2]
      c
      S[table-format=2.2]
      S[table-format=2.2]
      S[table-format=2.2]
      S[table-format=2.2]
    }
    \toprule
    & \multicolumn{7}{c}{LEGAR BENCH$_{\textit{Standard}}$ (P@5)} & \phantom{a} & \multicolumn{4}{c}{LEGAR BENCH$_{\textit{Strciter}}$ (P@5)} \\
    \cmidrule(lr){2-8} \cmidrule(lr){10-13}
    {Criminal Category} & {LegalSearchLM} & {BM25} & {Contriever} & {SAILER} & {Hybrid} & {mE5} & {OpenAI-Embedding} & & {LegalSearchLM} & {BM25} & {Contriever} & {SAILER} \\
    \midrule
    \multirow{1}{*}{Fraud}  & {0.74} & {0.53} & {0.57} & {0.64} & {0.66} & {0.63} & {0.65} &  & {0.33} & {0.33} & {0.03} & {0.24}  \\ \midrule 
    \multirow{1}{*}{Injury or Violence}  & {0.62} & {0.50} & {0.42} & {0.66} & {0.68} & {0.57} & {0.56} &  & {0.35} & {0.35} & {0.01} & {0.23} \\ 
    \midrule 
    \multirow{1}{*}{Sexual crime} & {0.62} & {0.49} & {0.40} & {0.65} & {0.66} & {0.51} & {0.53} &  & {0.37} & {0.37} & {0.02} & {0.32} \\ \midrule 
    \multirow{1}{*}{Finance or Insurance}  & {0.72} & {0.56} & {0.60} & {0.64} & {0.68} & {0.68} & {0.76} &  & {0.32} & {0.32} & {0} & {0.20} \\ 
    \midrule 
    \multirow{1}{*}{Defamation or Insult} & {0.83} & {0.58} & {0.48} & {0.78} & {0.78} & {0.68} & {0.70} &  & {0.33} & {0.33} & {0} & {0.22} \\ 
    \midrule 
    \multirow{1}{*}{Drug} & {0.80} & {0.52} & {0.76} & {0.84} & {0.84} & {0.88} & {0.84} &  & {0.34} & {0.33} & {0} & {0.10} \\ 
    \midrule 
    \multirow{1}{*}{Murder} & {0.50} & {0.50} & {0.30} & {0.50} & {0.50} & {0.30} & {0.30} &  & {0.39} & {0.39} & {0} & {0.35} \\ 
    \midrule     
    \multirow{1}{*}{Traffic offenses} & {0.47} & {0.33} & {0.33} & {0.94} & {0.94} & {0.82} & {0.89} &  & {0.34} & {0.34} & {0.02} & {0.12} \\ \midrule
    \multirow{1}{*}{\vdots} & {\vdots} & {\vdots} & {\vdots} & {\vdots} & {\vdots} & {\vdots} & {\vdots} &  & {} & {} & {} & {} \\ \midrule
    \multirow{1}{*}{\textbf{Total}} & {\textbf{0.68}} & {0.51} & {0.48} & {0.62} & {0.65} & {0.57} & {0.58} &  & {\textbf{0.35}} & {0.34} & {0.01} & {0.22} \\
    \bottomrule
    \end{tabular}
    }
    \caption{Results on LEGAR BENCH$_{\textit{Standard}}$ and LEGAR BENCH$_{\textit{Stricter}}$. We present results for the 8 crime categories shared by both evaluation sets, while the scores for the remaining 25 crime categories in LEGAR BENCH$_{\textit{Standard}}$ are listed in Table \ref{appendix:full_res}.}
    
    \label{tabs:standard_stricter_prec5}
    \end{threeparttable}
\end{table*}

\endgroup
\section{Experimental Setup}
\subsection{Baselines}
We evaluate a range of baselines, including traditional lexical matching and embedding-based methods trained on general or legal domains.

\paragraph{Lexical matching.} (1) \textsc{BM25}, a strong baseline in the legal domain~\cite{rosa2021yesbm25strongbaseline}, widely adopted by production-level legal RAG systems~\cite{magesh2024hallucinationfreeassessingreliabilityleading}.

\paragraph{General dual-encoders.} (1) \textsc{Contriever}~\cite{izacard2022unsuperviseddenseinformationretrieval}, a dual-encoder model designed as a general-purpose retriever. We further adapt this model by training it on our legal corpus. (2) \textsc{mE5}~\cite{wang2024multilinguale5textembeddings}, an open-source multilingual E5~\cite{wang2024textembeddingsweaklysupervisedcontrastive} embedding model that has achieved state-of-the-art retrieval performance. (3) \textsc{OpenAI-Embedding}~\footnote{\url{https://platform.openai.com/docs/guides/embeddings/embedding-models}}, a widely used commercial embedding model from OpenAI 
(\texttt{text-embedding-3-small}).

\paragraph{Legal dual-encoders.} (1) \textsc{Sailer}~\cite{li2023sailerstructureawarepretrainedlanguage}, which achieves strong performance in the LCR task of the COLIEE 2023 competition \citep{goebel2024overview}. Compared to \textsc{Contriever}, \textsc{Sailer} is pretrained using legal documents with section-level training loss (e.g., fact, interpretation (reasoning), and decision), enabling better legal document understanding. (2) \textsc{Hybrid}, obtained by averaging BM25 and \textsc{Sailer} scores.

\begin{figure}[t!]
{
\centering
    \begin{minipage}[b]{0.49\textwidth}
    \centering
    \includegraphics[width=\textwidth]{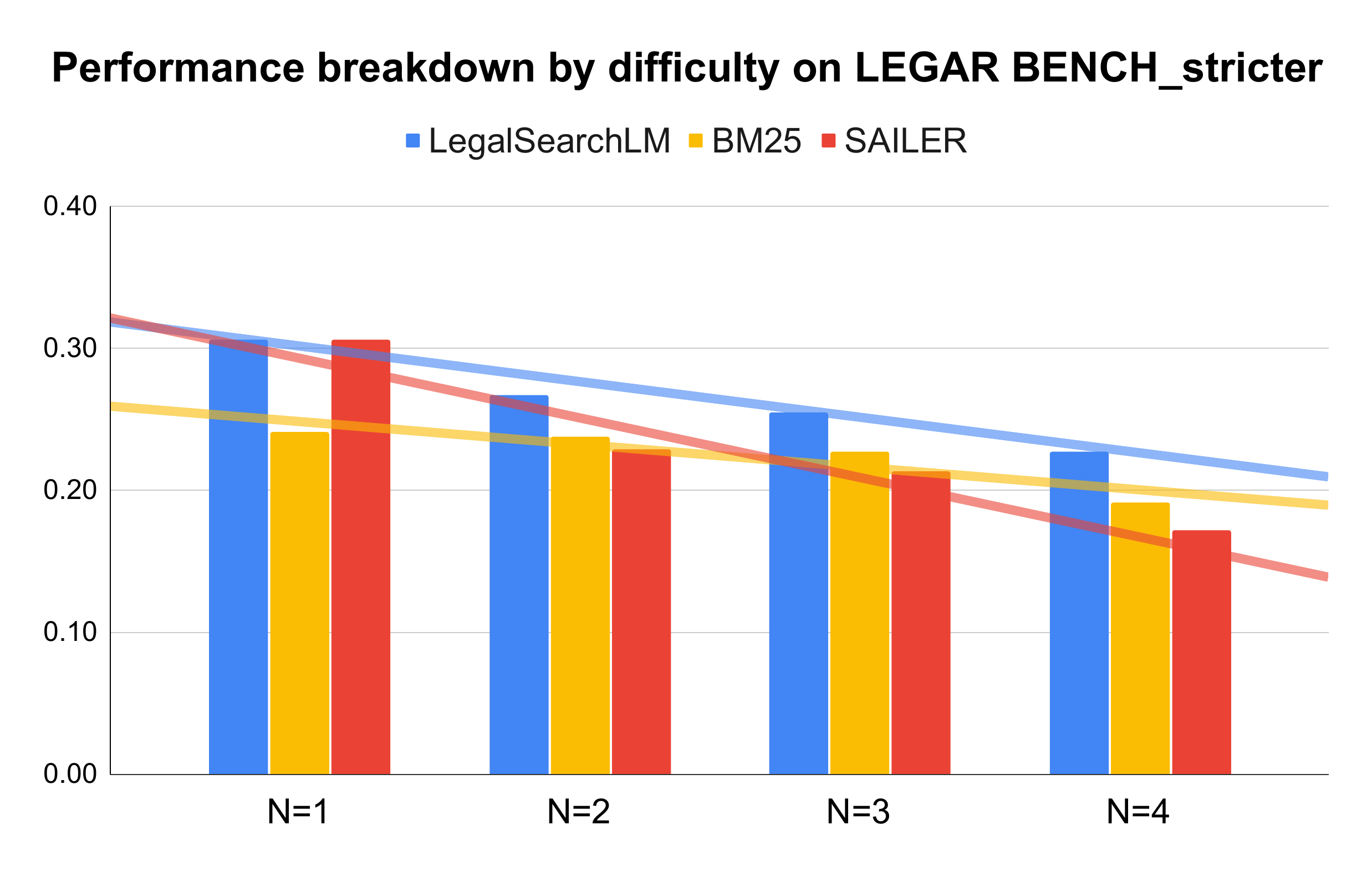}
    \caption{Performance on LEGAR BENCH$_{\textit{Stricter}}$ by four different difficulties, where $N$ represents the number of factors that should be matched. \textsc{LegalSearchLM} achieves the best performance across all difficulty levels, demonstrating robustness in complex retrieval settings.}
    \label{fig:breakdown_by_difficulty}
    \end{minipage}
}
\vspace{-20pt}
\end{figure}
\section{Results and Analysis}
\label{sec:experimental_results}
\paragraph{Performance on LEGAR BENCH$_{\textit{Standard}}$.} An evaluation on the standard version, consisting of 411 various query types across 33 crime categories, demonstrates that \textsc{LegalSearchLM} outperforms Contriever by 20\%, BM25 by 17\%, mE5 by 11\%, OpenAI-Embedding by 10\%, SAILER by 6\%, and Hybrid by 3\% (Table \ref{tabs:standard_stricter_prec5}, bottom row). Specifically, it outperformed BM25 in 28 crime categories, Contriever across all categories, and SAILER in 21 categories. In Table \ref{tabs:standard_stricter_prec5}, we provide the results for 8 out of 33 criminal categories for brevity. Full results are presented in Appendix \ref{appendix:full_res}, where we also provide a comparison with the reranked model, KELLER.

\paragraph{Performance on LEGAR BENCH$_{\textit{Stricter}}$.} An evaluation on the stricter version, which includes 15,777 diverse query types across 8 crime categories, further demonstrates \textsc{LegalSearchLM}'s effectiveness in handling complex legal knowledge, achieving the highest performance in Table \ref{tabs:standard_stricter_prec5}. BM25 excels at capturing fine-grained details through exact lexical matching, leading to stronger performance in LEGAR BENCH$_{\textit{Stricter}}$ compared to embedding-based similarity search. \textsc{LegalSearchLM} effectively captures both fine-grained details and legal semantic understanding, combining the strengths of both approaches. Further analysis is provided below.

\begin{figure}[t!]
{
\centering
    \begin{minipage}[b]{0.49\textwidth}
    \centering
    \includegraphics[width=\textwidth]{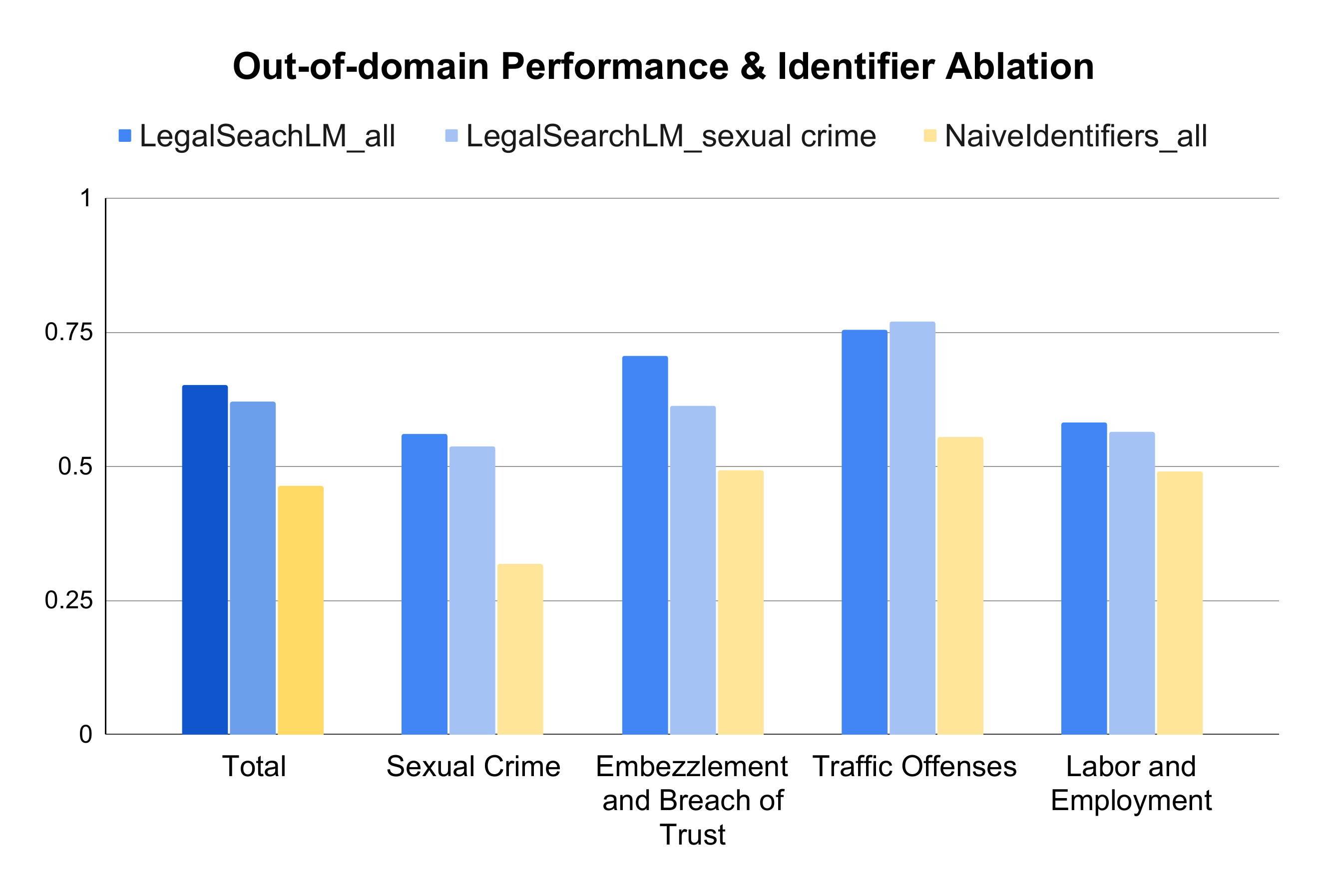}
    \caption{Performance on out-of-distribution. LegalSearchLM$_{\textit{all}}$ is trained on all test categories; LegalSearchLM$_{\textit{sexual crime}}$ on sexual crimes only. NaiveIdentifiers$_{\textit{all}}$ is trained with random spans within the query case as supervision. Performance of LegalSearchLM$_{\textit{sexual crime}}$ highlights its superior generalization ability.}
    \label{fig:out_of_distribution}
    \end{minipage}
}
\end{figure}

\label{sec:analysis_discussion}

\paragraph{Advantages of \textsc{LegalSearchLM} over existing retrieval methods.} 
In the LEGAL BENCH$_{\textit{Stricter}}$ setting, we analyze retrieval performance across varying levels of retrieval difficulty, measured by the number of sub-factors $N$.

As shown in Figure~\ref{fig:breakdown_by_difficulty}, The performance of SAILER, an embedding-based retriever, relatively sharply degrades in difficult problems compared to LegalSearchLM, indicating that information is lost when vectorized. On the other hand, BM25, a lexical matching method, demonstrates no significant change by difficulty since it retrieves in a way that captures overlapping keywords without regard to legal element understanding. In contrast, \textsc{LegalSearchLM} achieves the best performance in all difficulties by combining the strength of lexical matching in capturing fine-grained details with that of embedding-based retrieval in understanding semantics.

\paragraph{Generalization ability of \textsc{LegalSearchLM} to unseen crime types.} As legal professionals handle diverse cases, the generalizability to unseen criminal types is crucial in LCR. To evaluate this, we train \textsc{LegalSearchLM} only using sexual crime data and test it on unseen domains (embezzlement and breach of trust, traffic offenses, and labor and employment). We compare the results with a generative retrieval model trained on all crimes but with naive identifiers. Figure \ref{fig:out_of_distribution} shows that \textsc{LegalSearchLM} trained on sexual crime outperforms the model trained with naive identifiers by 15.66\%, despite the latter is trained using in-domain data. Furthermore, the performance is almost on par with \textsc{LegalSearchLM} trained on the full data. This demonstrates that the ability to effectively capture key legal elements is more beneficial than training on various datasets without carefully designed identifiers.

\section{Related Works}
\subsection{Legal Case Retrieval Datasets}
Legal Case Retrieval (LCR) is the task of retrieving target cases \textit{relevant} to a query case \citep{feng-etal-2024-legal, 10.1145/3404835.3463250}. While some works define relevant documents as one that is cited by the query \citep{shao2020thuircoliee2020leveragingsemanticunderstanding, li2023thuircoliee2023incorporatingstructural}, we focus on case similarity \citep{10.1145/3404835.3463250}.

In LCR, obtaining large-scale data is challenging as the annotation requires legal expertise. Therefore, existing works often restrict the crime types or the retrieval pool's size \citep{li2023lecardv2largescalechineselegal, 10.1145/3404835.3463250}. In contrast, this work successfully automated the annotation process while maintaining expert-level relevance judgments, eliminating the need for compromising data size and diversity.

In contrast, LEGAR BENCH$_{\textit{Standard}}$ has effectively scaled the number of distinct crimes and the number of documents in the retrieval pool by using statutory provisions. Furthermore, LEGAR BENCH$_{\textit{Stricter}}$ can evaluate the relevance based on expert-annotated legal factors on a large scale, which was not possible before.



\subsection{Legal Case Retrievers}

Earlier works on LCR have directly applied task-agnostic neural retrieval methods like cross-encoders \citep{xiao2021lawformerpretrainedlanguagemodel}. However, recent works emphasize the specificity of the legal domain. For instance, SAILER \citep{li2023sailerstructureawarepretrainedlanguage} incorporates the document structure of legal cases during the pretraining, improving the embedding quality. KELLER and Elem4LCR first segment the case into atomic legal elements, and apply element-wise embedding similarity (KELLER) or cross-encoder scoring (Elem4LCR) between cases to obtain a fine-grained similarity score. LegalSearchLM proposes novel LCR-specific adjustments specialized for generative language models, while previous works resort to encoder-only models.

\subsection{Generative Retrieval}
Generative Retrieval (GR) initially emerged from GENRE \cite{decao2021autoregressiveentityretrieval}, in which an encoder-decoder model retrieves a document by generating the title of the document from a given query.

Recent works explore identifiers based on specific document IDs~\cite{tay2022transformermemorydifferentiablesearch, mehta2023dsiupdatingtransformermemory, wang2023neuralcorpusindexerdocument, chen2024densexretrievalretrieval, zeng2023scalableeffectivegenerativeinformation, zeng2024planningaheadgenerativeretrieval} or the document's \textit{content}, which contain richer, finer-grained information. For instance, SEAL uses spans from the body text \cite{bevilacqua2022autoregressivesearchenginesgenerating}, which was also adopted by MINDER \citep{li-etal-2023-multiview} and LTRGR \citep{li2023learningrankgenerativeretrieval} that combine spans, titles, and pseudo-queries. \textsc{LegalSearchLM} is inspired by content identifiers and incorporates domain-specific adaptations to better address legal retrieval tasks. 




\section{Conclusion}
Legal case retrieval is a crucial task for legal practitioners, but resource scarcity and complex relevance judgment have hindered its application.
To address these issues, we first construct LEGAR BENCH, a novel large-scale LCR dataset based on expert-defined relevance. \textsc{LEGAR BENCH} comes with two subsets, one with broader crime coverage and one with finer-grained relevance labels, supporting diverse applications.
Next, we present \textsc{LegalSearchLM}, the first generative retrieval model, that captures the core legal elements relevant to the given query case. \textsc{LegalSearchLM} shows promising results in both versions of \textsc{LEGAR BENCH}, proving the potential of generative retrieval in LCR.

\section{Limitations}

In this dataset, we construct the largest benchmark in the legal case retrieval task, LEGAR BENCH. However, this dataset is restricted to the cases and statutes from the Korean legal system, which might limit its applicability beyond other jurisdictions and to non-Korean speakers. Furthermore, although legal experts were actively involved in defining the relevance criteria in LEGAR BENCH, they were not involved in the data point-wise verification of case-to-case relevance. As a result, there may be undetected noise in the dataset.

\section*{Acknowledgments}
We are grateful to LBOX for their support in constructing and releasing the benchmark datasets.


\bibliography{custom}

\newpage
\appendix

\clearpage
\section{Inference Details}
\subsection{Background on FM-index}
\label{appendix:fmindex}

\paragraph{FM-index.}
The FM-index is a compressed full-text index built on top of the Burrows--Wheeler Transform (BWT).
Given a text $T$ terminated by a unique sentinel (commonly ``\$''), 
the BWT is obtained by generating all cyclic rotations of $T$, sorting them lexicographically, 
and extracting the last column of the resulting matrix.

\paragraph{Data structures.}
The FM-index augments the BWT with two auxiliary structures. 
The array $C$ stores, for each character $c$, the number of characters in $T$ that are lexicographically smaller than $c$. 
The table $\mathrm{Occ}(c,i)$ counts how many occurrences of $c$ appear in the BWT up to position $i$. 
Together, these enable efficient navigation between the first and last columns of the BWT.

\paragraph{Backward search.}
Pattern matching proceeds right-to-left: we start with the entire text as a candidate range 
and, at each step, use $C$ and $\mathrm{Occ}$ to restrict the range to suffixes consistent with the processed part of the pattern. 
If the range becomes empty, the pattern does not occur; otherwise, the final range corresponds to all matches. 
This procedure runs in time proportional to the pattern length.

\paragraph{Use in constrained decoding.}
In our setting, the FM-index acts as a prefix-tree–like constraint over the corpus, 
restricting decoding to continuations that appear exactly in $T$. 
As a result, early token choices are critical, since they determine whether a valid continuation path remains available.

\section{Result Details}
\subsection{\texorpdfstring{Full results on LEGAR BENCH$_{\textit{Standard}}$}{Full results on LEGAR BENCH (Standard)}}

We provide the complete results on LEGAR BENCH$_{\textit{Standard}}$ across all 33 criminal categories in Table \ref{tabs:full_results_standard}. Keep in mind that KELLER, which focuses on reranking, leverages passage-level retrieval and make multiple inferences per case using majority voting (\textit{MaxSum}). This setup differs from our model and other baselines, making direct comparisons difficult. We include KELLER as a reference of a reranked model. All results are from single-run experiments.

\begin{table*}[t!]
    \centering
    \scriptsize
    \renewcommand{\arraystretch}{1.1}
    \sisetup{table-format=1.2, round-mode=places, round-precision=2}
    \begin{tabular}{
      l
      c
      c
      c
      c
      c
      c
      c
      c
      c
    }
    \toprule
    & \multicolumn{8}{c}{LEGAR BENCH$_{\textit{Standard}}$ (Precision@5)} \\
    \cmidrule(lr){2-9}
    {Criminal Category} & {\textbf{LegalSeachLM(Ours)}} & {BM25} & {Contriever} & {SAILER} & {\underline{Hybrid}} & {mE5} & {OpenAI-Embedding} & {KELLER*} \\
    \midrule
    \multirow{1}{*}{{[Total]}} & {\textbf{[0.68]}} & \cellcolor{cyan!30}{[0.51]} & {[0.48]} & {[0.62]} & {\underline{[0.65]}} & {[0.57]} & {[0.58]} & {[0.70]} \\ \midrule 
    \multirow{1}{*}{Traffic offenses} & {0.75} & {0.60} & {0.72} & {0.94} & {0.94} & {0.82} & {0.89} & {0.89} \\ \midrule 
    \multirow{1}{*}{Fraud} & {0.74} & {0.53} & {0.57} & {0.64} & {0.66} & {0.63} & {0.65} & {0.90} \\ \midrule 
    \multirow{1}{*}{Injury or Violence}  & {0.62} & {0.50} & {0.42} & {0.66} & {0.68} & {0.57} & {0.56} & {0.70} \\ \midrule 
    \multirow{1}{*}{Sexual crime} & {0.52} & {0.49} & {0.40} & {0.65} & {0.66} & {0.51} & {0.53} & {0.73} \\ \midrule 
    \multirow{1}{*}{Theft or Robbery}  & {0.65} & {0.48} & {0.41} & {0.60} & {0.63} & {0.48} & {0.47} & {0.69} \\ \midrule 
    \multirow{1}{*}{Obstruction of Business}  & {0.75} & {0.58} & {0.40} & {0.74} & {0.74} & {0.65} & {0.57} & {0.94} \\ \midrule 
    \multirow{1}{*}{Embezzlement or Breach of trust} & {0.67} & {0.64} & {0.49} & {0.56} & {0.60} & {0.55} & {0.60} & {0.81} \\ \midrule 
    \multirow{1}{*}{Destruction}  & {0.64} & {0.64} & {0.48} & {0.72} & {0.68} & {0.48} & {0.60} & {0.92} \\ \midrule 
    \multirow{1}{*}{Finance or Insurance} & {0.72} & {0.56} & {0.60} & {0.64} & {0.68} & {0.68} & {0.76} & {0.92} \\ \midrule 
    \multirow{1}{*}{Threat} & {0.60} & {0.60} & {0.47} & {0.60} & {0.64} & {0.62} & {0.75} & {0.87} \\ \midrule 
    \multirow{1}{*}{Defamation or Insult} & {0.83} & {0.58} & {0.48} & {0.78} & {0.78} & {0.68} & {0.70} & {0.80} \\ \midrule 
    \multirow{1}{*}{Drug} & {0.80} & {0.52} & {0.76} & {0.84} & {0.84} & {0.88} & {0.84} & {0.80} \\ \midrule 
    \multirow{1}{*}{Criminal trespass} & {0.73} & {0.63} & {0.51} & {0.80} & {0.80} & {0.65} & {0.63} & {0.89} \\ \midrule 
    \multirow{1}{*}{Gambling} & {0.74} & {0.54} & {0.63} & {0.71} & {0.74} & {0.89} & {0.74} & {0.97} \\ \midrule 
    \multirow{1}{*}{Negligent homicide and injury} & {0.37} & {0.27} & {0.30} & {0.57} & {0.57} & {0.43} & {0.40} & {0.50} \\ \midrule 
    \multirow{1}{*}{Obstruction of right} & {0.80} & {0.52} & {0.52} & {0.52} & {0.56} & {0.40} & {0.52} & {0.92} \\ \midrule 
    \multirow{1}{*}{Child abuse or School violence} & {0.64} & {0.48} & {0.38} & {0.50} & {0.52} & {0.62} & {0.50} & {0.54} \\ \midrule 
    \multirow{1}{*}{Medical or Food drug} & {0.40} & {0.35} & {0.22} & {0.11} & {0.24} & {0.25} & {0.22} & {0.14} \\ \midrule  
    \multirow{1}{*}{Murder} & {0.50} & {0.50} & {0.30} & {0.50} & {0.50} & {0.30} & {0.30} & {0.60} \\ \midrule     \multirow{1}{*}{Corporation} & {0.47} & {0.33} & {0.33} & {0.33} & {0.47} & {0.33} & {0.27} & {0.40} \\ \midrule 
    \multirow{1}{*}{Bribery} & {0.47} & {0.60} & {0.40} & {0.80} & {0.80} & {0.47} & {0.53} & {0.80} \\ \midrule 
    \multirow{1}{*}{Car} & {0.80} & {0.60} & {0.60} & {0.60} & {0.60} & {0.60} & {0.70} & {1.00} \\ \midrule 
    \multirow{1}{*}{Labor or Employment} & {0.53} & {0.51} & {0.40} & {0.58} & {0.64} & {0.56} & {0.53} & {0.55} \\ \midrule 
    \multirow{1}{*}{Industrial or Serious accidents} & {0.45} & {0.45} & {0.20} & {0.40} & {0.50} & {0.50} & {0.30} & {0.25} \\ \midrule 
    \multirow{1}{*}{Military duty or law} & {0.70} & {0.50} & {0.50} & {0.60} & {0.60} & {0.60} & {0.60} & {0.50} \\ \midrule
    \multirow{1}{*}{Consumer or Fair trade} & {1.00} & {0.60} & {0.80} & {0.40} & {0.40} & {0.60} & {0.40} & {0.40} \\ \midrule 
    \multirow{1}{*}{Arrest or Detention} & {1.00} & {0.80} & {0.40} & {0.80} & {0.80} & {0.40} & {0.60} & {1.00} \\ \midrule 
    \multirow{1}{*}{Intellectual property} & {0.60} & {0.33} & {0.67} & {0.80} & {0.80} & {0.80} & {0.73} & {0.73} \\ \midrule 
    \multirow{1}{*}{IT or Privacy} & {1.00} & {0.60} & {0.80} & {1.00} & {1.00} & {0.80} & {0.60} & {1.00} \\ \midrule 
    \multirow{1}{*}{Misdemeanor} & {0.40} & {0.20} & {0.20} & {0.40} & {0.40} & {0.40} & {0.20} & {0.40} \\ \midrule 
    \multirow{1}{*}{Sexual norms} & {0.80} & {0.20} & {0.20} & {0.20} & {0.20} & {0.40} & {1.00} & {0} \\ \midrule 
    \multirow{1}{*}{Tax, Administ, Const law} & {0.81} & {0.61} & {0.71} & {0.91} & {0.91} & {0.74} & {0.71} & {0.76} \\ \midrule 
    \multirow{1}{*}{Other criminal offenses} & {0.72} & {0.58} & {0.42} & {0.70} & {0.74} & {0.66} & {0.70} & {0.88} \\
    \bottomrule
    \end{tabular}
    \caption{Full results of LegalSearchLM and the baselines on LEGAR BENCH$_{\text{Standard}}$ across all 33 criminal categories.}
    \label{tabs:full_results_standard}
\end{table*}

\label{appendix:full_res}
\section{Benchmark Details}








\subsection{Collaboration with legal experts}
\label{appendix:annotation}
Annotation of LEGAR BENCH$_{\textit{Stricter}}$ requires a significant amount of legal expert annotation. For instance, determining the critical factors that determine the applicability of a specific statute requires extensive knowledge of criminal law, and determining the range of inherently continuous values (\textit{e.g.} severity of an injury) that are similarly treated in practice requires strong expertise in practicing criminal law.

For expert annotation, we hired five Korean lawyers specialized in the Criminal Act to construct our LEGAR BENCH$_{\textit{Standard}}$ and LEGAR BENCH$_{\textit{Stricter}}$. The lawyers were instructed to organize high-coverage categories that encompass most criminal offenses and to label subcategories of cases based on charge titles. Additionally, for the stricter version, they listed relevant factors (factual details) for each specific charge in LEGAR BENCH$_{\textit{Standard}}$. Lawyers spent a total 70\ hours for the annotation task, and the compensation was approximately \$250/hour during the whole process.

\subsection{\texorpdfstring{Statistics of LEGAR BENCH$_{\textit{Standard}}$}{Statistics of LEGAR BENCH (Standard)}}

\begin{table}[t!]
    \centering
    \small
    \fontsize{12}{17}\selectfont
    \resizebox{0.48\textwidth}{!}{
        \begin{tabular}{cccc}
            \toprule
             Crime categories & \# of Standard group & \# of Cases & \\
             \midrule
            Traffic offenses & 13 & 319,527 \\
            Fraud & 21 & 181,703 \\
            Injury or Violence & 31 & 146,764 \\
            Sexual crime & 132 & 104,919 \\
            Theft or Robbery & 38 & 74,772 \\
            Obstruction of Business & 13 & 74,722 \\
            Embezzlement or Breach of trust & 15 & 39,835 \\
            Destruction & 5 & 39,595 \\
            Finance or Insurance & 5 & 32,944 \\
            Threat & 11 & 27,496 \\
            Defamation or Insult & 8 & 27,278 \\
            Drug & 5 & 26,066 \\
            Criminal trespass & 15 & 24,856 \\
            Gambling & 7 & 11,091 \\
            Negligent homicide and injury & 6 & 7,384 \\
            Obstruction of right & 5 & 6,749 \\
            Child abuse or School violence & 10 & 5,756 \\
            Medical or Food drug & 11 & 98 \\
            Murder & 2 & 4,306 \\
            Corporation & 3 & 1,195 \\
            Bribery & 3 & 1,638 \\
            Car & 2 & 20,882 \\
            Labor or Employment & 11 & 12,647 \\
            Industrial or Serious accidents & 4 & 198 \\
            Military duty or law & 2 & 9,300  \\
            Consumer or Fair trade & 1 & 128 \\            
            Arrest or Detention & 1 & 6 \\     
            Intellectual property & 3 & 3,927  \\
            IT or Privacy & 2 & 2,311 \\
            Misdemeanor & 1 & 6,476 \\
            Sexual norms & 1 & 4,140 \\
            Tax, Administ, Const law & 14 & 40,890 \\
            Other criminal offenses & 10 & 23,211 \\
            \midrule
            \textbf{Total} & \textbf{411} & \textbf{1,052,506} \\
            \bottomrule
        \end{tabular}
    }
    \caption{Statistics of Crime typology and Standard version of LEGAR BENCH. The total number of cases is reported as a unique count, excluding duplicates from cases classified under multiple categories \(1,347,962 \rightarrow 1,052,506\).}
    \label{tables:standard_stats}
\end{table}
Table \ref{tables:standard_stats} presents a criminal typology that includes 33 major categories of criminal offenses. Each category is classified in detail based on charge titles and statutes. The number of standard groups for each category is listed under \textit{\#of Standard Group}, while the number of unique case documents mapped to each group is listed under \textit{\#of Cases}. The total number of standard groups is 411, including 1,052,506 unique cases, which constitute 85.79\% of the entire corpus (1,226,814 cases). This figure underscores the broad coverage of our benchmark across a wide range of types of criminal offenses.

\subsection{\texorpdfstring{Statistics of LEGAR BENCH$_{\textit{Stricter}}$}{Statistics of LEGAR BENCH (Stricter)}}
Table \ref{tables:stricter_stats} shows statistics of stricter groups in 8 criminal categories. The number of stricter groups for each category is listed under \textit{\#of Stricter Group}, while the number of unique case documents mapped to each group is listed under \textit{\#of Cases}.

\begin{table}[t!]
    \centering
    \small
    \fontsize{9}{14}\selectfont
    \resizebox{0.48\textwidth}{!}{
        \begin{tabular}{cccc}
            \toprule
             Crime categories & \# of Stricter group & \# of Cases & \\
             \midrule
            Fraud & 8 & 325 \\
            Injury or Violence & 19 & 308 \\
            Sexual crime & 111 & 1,061 \\
            Finance or Insurance & 1 & 28 \\
            Defamation or Insult & 6 & 253 \\
            Drug & 4 & 37 \\
            Murder & 2 & 8 \\
            Traffic offenses & 9 & 330 \\
            \midrule
            \textbf{Total} & \textbf{160} & \textbf{2,350} \\
            \bottomrule
        \end{tabular}
    }
    \caption{Statistics of Stricter version of LEGAR BENCH.}
    \label{tables:stricter_stats}
\end{table}

\subsection{Total crime types in 33 categories for query cases.}
\paragraph{Traffic offenses.}
In Table \ref{tab:korean_traffic}.
\paragraph{Sexual crime.}
In Table \ref{tab:korean_sexual_crime}
\paragraph{Fraud.}
In Table \ref{tab:korean_fraud}.
\paragraph{Injury and Violence.}
In Table \ref{tab:korean_violence}.
\paragraph{Theft and Robbery.}
In Table \ref{tab:korean_theft_robbery}.
\paragraph{Embezzlement.}
In Table \ref{tab:korean_embezzlement}.
\paragraph{Destruction.}
In Table \ref{tab:korean_destruction}.
\paragraph{Finance and Insurance.}
In Table \ref{tab:korean_finance_insurance}.
\paragraph{Threat.}
In Table \ref{tab:korean_threat}.
\paragraph{Crimes against Reputation.}
In Table \ref{tab:korean_defamation}.
\paragraph{Drug.}
In Table \ref{tab:korean_drug}.
\paragraph{Gambling.}
In Table \ref{tab:korean_gambling}.
\paragraph{Negligent homicide and injury.}
In Table \ref{tab:korean_negligent}.
\paragraph{Obstruction of rights.}
In Table \ref{tab:korean_obs_right}.
\paragraph{Crimes against children and School violence.}
In Table \ref{tab:korean_child_school}.
\paragraph{Medical and Food drug.}
In Table \ref{tab:korean_medical_food_drug}.
\paragraph{Murder.}
In Table \ref{tab:korean_murder}.
\paragraph{Corporation.}
In Table \ref{tab:korean_corporation}.
\paragraph{Bribery.}
In Table \ref{tab:korean_bribery}.
\paragraph{Labor and Employment.}
In Table \ref{tab:korean_labor}.
\paragraph{Fair trade.}
In Table \ref{tab:korean_fair_trade}.
\paragraph{Arrest and Detention.}
In Table \ref{tab:korean_arrest}.
\paragraph{Other criminal offenses.}
In Table \ref{tab:korean_other_crime}.
\paragraph{Car-related offenses.}
In Table \ref{tab:korean_car}.
\paragraph{Home invasion.}
In Table \ref{tab:korean_home_invasion}.
\paragraph{Industrial accident or Serious accident.}
In Table \ref{tab:korean_industry_serious}.
\paragraph{Intellectual property rights.}
In Table \ref{tab:korean_intellecual}.
\paragraph{IT or Privacy.}
In Table \ref{tab:korean_it_privacy}.
\paragraph{Military duty.}
In Table \ref{tab:korean_military_duty}.
\paragraph{Misdemeanor.}
In Table \ref{tab:korean_misdemeanor}.
\paragraph{Obstruction of business.}
In Table \ref{tab:korean_obs_business}.
\paragraph{Sexual morality.}
In Table \ref{tab:korean_sexual_morality}.
\paragraph{Tax or Administrative or Constitutional Law.}
In Table \ref{tab:korean_tax}.

\subsection{Full list of the stricter relevance group}

LEGAR BENCH$_{\textit{Stricter}}$ further divides LEGAR BENCH$_{\textit{Standard}}$ categories based on different factual details of a criminal case that do not affect the type of charge, but might affect the final judgment (guilty or innocent) or the sentence \textit{e.g.} information about defendant/victims, methods, consequences, and claims made in court. Also, it provides a comprehensive list of possible options for each factor.
The options are primarily based on the official sentencing guidelines from the Sentencing Commission of the Supreme Court of Korea, and annual crime statistics reports published by government/academic authorities including the Supreme Prosecutor's Office and the Korean Institute of Criminology. However, these lists are often insufficient to express existing cases, especially the defendant's claims (\textit{e.g.}, a defendant convicted of assault might claim that the act was due to self-defense, pleading for innocence). Identifying such factors heavily relies on deep understanding and expertise in practicing law. Hence, the lawyers were instructed to add factors and options that are frequent and important in practice but not mentioned in the official documents. 

Previous work in identifying such factors in the Korean Criminal Act \citet{hwang-etal-2022-data} includes only 11 unique factors across 4 crime categories focusing only on facts, while this work adds 102 unique factors (including 39 defendant claims) across 8 categories.

\subsection{Prompt template for LEGAR BENCH\texorpdfstring{$_{\textit{Stricter}}$}{ (Stricter)} annotation}

The example annotation templates are shown on the following page. The blue box presents the original Korean version, and the pink box presents the English version provided for reference.


\begin{figure*}[p]
\begin{tcolorbox}[colback=white!90!cyan!10, colframe=cyan!50!white,
                  title=Example prompt for a specific d1--d2 case (Korean)]
\tiny
Input으로 주어지는 판결문은 {d1} 카테고리의 {d2} 범죄에 대한 내용입니다. 아래 각 항목의 분류 기준을 기반으로 항목별 분류 번호를 제공하세요. 모든 항목에 대해 누락없이 분류 번호를 제공하세요. 단, 답변은 Output 예시처럼 항목과 분류 번호로 구성된 튜플이 나열된 리스트 형식으로 작성하세요.
\\
\\
{"d1": "명예훼손및모욕",\\
"d2": "허위적시명예훼손",\\
"description":\\
"<형사공통>\\
- 항목: 자수여부\\
- 항목 설명: 피고인이 범행 이후 자수하였는지 여부\\
- 분류 기준:\\
1. 피고인이 자수함 
2. 피고인이 자수하지 않음 
\\\\
- 항목: 심신미약여부\\
- 항목 설명: 피고인이 범행 당시 심신미약이었는지 여부\\
- 분류 기준:\\
1. 해당 쟁점이 다루어지고, 인정됨 (사실관계에 '심신미약 상태로...'의 표현이 있는 경우 포함) 
2. 해당 쟁점이 다루어지고, 인정되지 않음 
3. 해당 쟁점이 다뤄지지 않음 
\\\\
- 항목: 심신상실여부\\
- 항목 설명: 피고인이 범행 당시 심신상실이었는지 여부\\
- 분류 기준:\\ 1. 해당 쟁점이 다루어지고, 인정됨 (사실관계에 '심신상실 상태로...'의 표현이 있는 경우 포함) 
2. 해당 쟁점이 다루어지고, 인정되지 않음 
3. 해당 쟁점이 다뤄지지 않음 
\\\\
<피해자범죄>\\
- 항목: 피해자수\\
- 항목 설명: 피해자의 수\\
- 분류 기준:\\
1. 1명 
2. 2명 
3. 3명 이상 
\\\\
- 항목: 피해자와의관계\\
- 항목 설명: 피고인과 피해자와의 관계\\
- 분류 기준:\\
1. 연인 
2. 부부 
3. 피해자가 피고인의 부모 
4. 피해자가 피고인의 자녀 
5. 5촌 이내 친족 
6. 친구/지인 
7. 피해자가 피고인의 피감독자/피보호자 (교사/제자, 고용인/피고용인 등) 
8. 동료직원 
9. 클럽, 랜덤채팅 등 일회성 만남 
10. 불특정 다수 대상 범행 (피해자와 가해자가 모르는 사이인 경우) 
\\\\
<명예훼손>\\
- 항목: 명예훼손유형\\
- 항목 설명: 명예훼손의 문제가 된 발언의 유형\\
- 분류 기준:\\
1. 과거 전과 기록 
2. 직업적 명예와 관한 사실 (비리, 중대한 업무상 과실, 업무능력에 대한 평가 등) 
3. 범죄행위 등과 관련된 의혹 제기 
4. 치정/가족사/사생활 등 
5. 기타 
\\\\
- 항목: 명예훼손매체유형\\
- 항목 설명: 명예훼손의 매체\\
- 분류 기준:\\
1. 일상적인 개인 간 대화 
2. 연설/토론회 등 공적 대화 
3. 벽보, 포스터, 공고문 등 공개된 문서 
4. 정보통신망-인터넷 뉴스 등 공적 매체 
5. 정보통신망-SNS, 커뮤니티, 채팅, 게임, 유튜브 등 개인 간의 대화 
6. 출판물-신문, 잡지 등 
7. 출판물-서적, 논문 등 
8. 출판물-TV/라디오 등 매체 
\\\\
- 항목: 피해자유형\\
- 항목 설명: 피해자가 자연인인지 법인인지 여부\\
- 분류 기준:\\
1. 자연인 
2. 법인 
3. 법인격없는단체 (공중 등) 
\\\\
- 항목: 구체적사실의적시\\
- 항목 설명: 명예훼손죄가 성립하기 위한 구체적인 사실이 적시되었는지 여부가 쟁점인 경우\\
- 분류 기준:\\
1. 해당 쟁점이 다루어지고, 인정됨 
2. 해당 쟁점이 다루어지고, 인정되지 않음 
3. 해당 쟁점이 다뤄지지 않음 
\\\\
- 항목: 공연성전파가능성\\
- 항목 설명: 불특정 또는 다수인이 인식할 수 있는 상태 (공연성) 를 만족하는지, 제3자에게 전파될 가능성이 있는지가 쟁점인 경우\\
- 분류 기준:\\
1. 해당 쟁점이 다루어지고, 인정됨 
2. 해당 쟁점이 다루어지고, 인정되지 않음 
3. 해당 쟁점이 다뤄지지 않음 
\\\\
- 항목: 피해자특정성\\
- 항목 설명: 명예훼손만의 내용으로 피해자가 특정되는지 여부\\
- 분류 기준:\\
1. 해당 쟁점이 다루어지고, 인정됨 
2. 해당 쟁점이 다루어지고, 인정되지 않음 
3. 해당 쟁점이 다뤄지지 않음 
\\\\
- 항목: 사회적평가저해여부\\
- 항목 설명: 명예훼손의 내용이 피해자의 명예를 훼손하여 사회적 평가의 저해로 이어지는 사안인지가 쟁점이 된 경우\\
- 분류 기준:\\
1. 해당 쟁점이 다루어지고, 인정됨 
2. 해당 쟁점이 다루어지고, 인정되지 않음 
3. 해당 쟁점이 다뤄지지 않음 
\\\\
- 항목: 공익위법성조각정당행위\\
- 항목 설명: 명예훼손의 내용이 진실한 사실로서 오직 공공의 이익을 목적으로 하므로 위법성이 조각되거나, 언론사 등의 적법한 업무로서 정당한 행위인지 여부가 쟁점이 되는 경우\\
- 분류 기준:\\
1. 해당 쟁점이 다루어지고, 인정됨 
2. 해당 쟁점이 다루어지고, 인정되지 않음 
3. 해당 쟁점이 다뤄지지 않음"
\\\\
Output 예시:
[('피해자와의관계', 6), ('명예훼손유형', 2), ('명예훼손의 매체', 5), ('공연성전파가능성', 1), ('공익위법성조각정당행위', 3)]
\\\\
Input: \#d1카테고리 (명예훼손및모욕) d2범죄 (허위적시명예훼손)에 해당되는 각 판결문 예시 (생략)}

\end{tcolorbox}
\label{fig:stricter_llm_annotation_korean}
\end{figure*}

\begin{figure*}[p]
\begin{tcolorbox}[colback=pink!5, colframe=pink!60!red!80,
                  title=Example prompt for a specific d1--d2 case (English)]
\tiny
The judgment provided as input concerns an offense in category {d1} and specific crime type {d2}. Based on the classification criteria for each item below, provide the classification number for each item. Provide a classification number for every item without omission. The answer must be written in a list format consisting of tuples of (item, classification number), as in the Output example.
\\
\\
{"d1": "Defamation and Insult",\\
"d2": "Defamation by False Facts",\\
"description":\\
<Criminal (Common)>\\
- Item: Voluntary surrender\\
- Item description: Whether the defendant voluntarily surrendered after the offense\\
- Classification criteria:\\
1. The defendant voluntarily surrendered 
2. The defendant did not voluntarily surrender 
\\\\
- Item: Diminished capacity\\
- Item description: Whether the defendant was in a state of diminished capacity at the time of the offense\\
- Classification criteria:\\
1. The issue is addressed and recognized (including cases where the facts state 'in a state of diminished capacity...') 
2. The issue is addressed but not recognized 
3. The issue is not addressed 
\\\\
- Item: Insanity\\
- Item description: Whether the defendant was insane at the time of the offense\\
- Classification criteria:\\ 1. The issue is addressed and recognized (including cases where the facts state 'in a state of insanity...') 
2. The issue is addressed but not recognized 
3. The issue is not addressed 
\\\\
<Victim-Related Offense>\\
- Item: Number of victims\\
- Item description: Number of victims\\
- Classification criteria:\\
1. 1 victim 
2. 2 victims 
3. 3 or more victims 
\\\\
- Item: Relationship with the victim\\
- Item description: Relationship between the defendant and the victim\\
- Classification criteria:\\
1. Romantic partner 
2. Spouses 
3. The victim is the defendant's parent 
4. The victim is the defendant's child 
5. Relative within the fifth degree of kinship 
6. Friend/acquaintance 
7. The victim is the defendant's supervisee/ward (teacher/student, employer/employee, etc.) 
8. Coworker 
9. One-off encounter (club, random chat, etc.) 
10. Offense against unspecified persons (the perpetrator and victim are strangers) 
\\\\
<Defamation>\\
- Item: Type of defamatory statement\\
- Item description: The type of statement at issue for defamation\\
- Classification criteria:\\
1. Prior criminal record 
2. Facts concerning professional reputation (corruption, serious professional negligence, evaluation of job ability, etc.) 
3. Allegations related to criminal conduct, etc. 
4. Romantic/family/private life, etc. 
5. Other 
\\\\
- Item: Medium of defamation\\
- Item description: The medium of the defamatory act\\
- Classification criteria:\\
1. Everyday private conversation 
2. Public speech/discussion (e.g., lecture, debate) 
3. Publicly posted documents (posters, notices, etc.) 
4. Information network – internet news and other public media 
5. Information network – interpersonal channels such as SNS, communities, chat, games, YouTube, etc. 
6. Publications – newspapers, magazines, etc. 
7. Publications – books, academic papers, etc. 
8. Publications – TV/radio, etc. 
\\\\
- Item: Type of victim\\
- Item description: Whether the victim is a natural person or a legal entity\\
- Classification criteria:\\
1. Natural person 
2. Legal entity 
3. Unincorporated association (the public, etc.) 
\\\\
- Item: Statement of specific facts\\
- Item description: When it is at issue whether specific facts were stated, which is required for defamation to be established\\
- Classification criteria:\\
1. The issue is addressed and recognized 
2. The issue is addressed but not recognized 
3. The issue is not addressed 
\\\\
- Item: Publicity / possibility of dissemination\\
- Item description: When it is at issue whether the requirement of publicity (recognizable by unspecified or many persons) is satisfied and whether there is a possibility of dissemination to third parties\\
- Classification criteria:\\
1. The issue is addressed and recognized 
2. The issue is addressed but not recognized 
3. The issue is not addressed 
\\\\
- Item: Specificity of the victim\\
- Item description: Whether the victim is identifiable solely from the content of the defamatory statement\\
- Classification criteria:\\
1. The issue is addressed and recognized 
2. The issue is addressed but not recognized 
3. The issue is not addressed 
\\\\
- Item: Impairment of social reputation\\
- Item description: When it is at issue whether the content of the defamation harms the victim's reputation and leads to impairment of social evaluation\\
- Classification criteria:\\
1. The issue is addressed and recognized 
2. The issue is addressed but not recognized 
3. The issue is not addressed 
\\\\
- Item: Public-interest / justifiable act (illegality exemption)\\
- Item description: When it is at issue whether the content of the defamation consists of true facts for the sole purpose of the public interest, thereby excluding illegality, or constitutes a justifiable act as part of legitimate activities such as those of the press\\
- Classification criteria:\\
1. The issue is addressed and recognized 
2. The issue is addressed but not recognized 
3. The issue is not addressed 
\\\\
Output example:
[('Relationship with the victim', 6), ('Type of defamatory statement', 2), ('Medium of defamation', 5), ('Publicity / possibility of dissemination', 1), ('Public-interest / justifiable act (illegality exemption)', 3)]
\\\\
Input: (omitted) example case corresponding to each d1-d2 pair}

\end{tcolorbox}
\label{fig:stricter_llm_annotation_english}
\end{figure*}


\begin{table*}[ht]
    \centering
    \small
    \begin{tabular}{p{3cm}p{9.5cm}}
        \toprule
        Crime categories & Factors(\# Options)\\

        \midrule
        Traffic offenses & Traffic accident type(6), Traffic accident time(2), Automobile type(3), Road type(4), Gross negligence type(18), Automobile accident insurance(3), Malpractice?(3), Hit-and-run type(3), Hit-and-run loss type(2), Aided victim?(3), Not aware of accident?(3), Blood alcohol level(3), Driving distance(4), Necessity?(3), Not driving?(3), Absorption phase?(3), Excessive extrapolation?(3), Driving without license type(5), Not aware of license suspension(3), Not aware of invalidation(3), Injury severity(8), Injury?(3), Number of victims(3), Defendant-victim relation(10), Surrender(2), Defendant feeble-minded?(3), Defendant insanity?(3), Reason not reaching consummation(4), Reached consummation?(3)  \\ \midrule
        
        Fraud &  Fraud type(14), No intent for pecuniary advantage?(3), No intent to defraud?(3), Profit(12), Defendant feeble-minded?(3), Defendant insanity?(3)  \\ \midrule
        Injury or Violence &  Two-way assault(2), Motivation(7), Intent to injure?(3), Self-defense?(3), Assault method(9), Injury severity(8), Injury?(3), Special crime type(2), Number of accomplices(5), Dangerous weapon?(3), Time between injury and death(4), Injury direct cause of death?(3), Surrender(2), Defendant feeble-minded?(3), Defendant insanity?(3) \\ \midrule
        
        Sexual crime &  Sexual assault location(6), Victim age(4), Victim disability(2), Defendant under influence(3), Victim under influence(3), Consent?(3), Intercourse type(4), Incident act type(4), Incident act by blitz(2), Victim sexual shame(3), Inability to resist cause(5), Aware of inability to resist?(3), Aware of victim's age under 13?(3), Aware of victim's age under 16?(3), Fraudulence/influence type(7), Victim under influence?(3), Covert photography filming/distribution type(7), Number of covert photography(4), Profit(4), Obscene communication medium(4), Obscene communication content(6), Object of sexual satisfaction(2), Reached the victim?(3), Assault/threat type(6), Assault method(9), Injury severity(8), Injury?(3), Special crime type(2), Number of accomplices(5), Dangerous weapon?(3), Time between injury and death(4), Injury direct cause of death?(3), No intent to defraud?(3), Number of victims(3), Defendant-victim relation(10), Surrender(2), Defendant feeble-minded?(3), Defendant insanity?(3), Reason not reaching consummation(4), Reached consummation?(3)  \\ \midrule
        
        Finance or Insurance~\footnote{Only insurance fraud cases have stricter relevance annotations.} & Insurance fraud type(5), No intent for pecuniary advantage?(3), No intent to defraud?(3), Profit(12), Surrender(2), Defendant feeble-minded?(3), Defendant insanity?(3), Reason not reaching consummation(4), Reached consummation?(3)  \\ \midrule
        
        Defamation or Insult &  Defamation content(5), Defamation medium(8), Insult content(4), Victim type(3), Alleged facts?, Publicly alleged?(3), Can specify victim?(3), Defaming the social status?(3), Justified(3), Number of victims(3), Defendant-victim relation(10), Surrender(2), Defendant feeble-minded?(3), Defendant insanity?(3) \\ \midrule
        
        Drug &  Drug type(14), Drug crime type(7), Defendant role(6), Narcotic handling license(6), Drug quantity(6), Profit(12), Surrender(2), Defendant feeble-minded?(3), Defendant insanity?(3)  \\ \midrule
        
        Murder &  Motivation(7), Intent to kill?(3), Self-defense?(3), Assault method(9), Injury?(3), Number of victims(3), Defendant-victim relation(10), Surrender(2), Defendant feeble-minded?(3), Defendant insanity?(3), Reason not reaching consummation(4), Reached consummation?(3) \\
        \bottomrule
    \end{tabular}
    \caption{Factors for defining Stricter relevance. Each factor is presented with the number of options in parentheses. Question mark(?) indicates that the factor represents a \textit{claim} defendant makes in a court, which always has three options (not mention, claimed but not taken, claimed and taken). As some factors only apply to certain standard groups (\textit{e.g.} Traffic accident type(6) only applies to traffic crimes involving accidents and not crimes like Driving Under the Influence (without any traffic accident)) and not all combinations are possible (\textit{e.g.} Killing Ascendant (killing one’s own or any lineal ascendant of one’s spouse) cases can only take two options (\textit{parent, other family members}) out of 10 options (\textit{partners, friend, ...}) provided for the Defendant-victim relation factor), the total number of stricter groups is a magnitude smaller compared to all option numbers multiplied.}
    \label{tab:stricter_list}
\end{table*}


\section{Implementation Details}
All models are trained using 8 * A100 80GB GPUs.

\paragraph{LegalSearchLM.}
To develop our SearchLM based on an autoregressive language model, we take the mt5-base pretrained model \citep{xue-etal-2021-mt5} and train it on 170K cases for a single epoch. We create a training dataset with a maximum of 15 query case-element pairs and 5 element-element pairs.

\paragraph{Contriever.}
We select Contriever as a representative model for retrieval in the general domain. We perform unsupervised training on the bert-base-multilingual-cased pretrained model with 170K cases for 10 epochs. Following the results in their work, we use the MoCo method during training rather than in-batch.

\paragraph{SAILER.}
We implement SAILER as a representative model for retrieval in the legal domain. Following their paper, we pretrain the bert-base-multilingual-cased model on facts, interpretations, and decisions of 1.2M cases for a single epoch, using the same configuration as in SAILER. The pretrained model is then fine-tuned for a single epoch with positive and negative samples, adjusting the learning rate from the default 5e-6 to 5e-5. We retrieve 100 related cases using BM25 over the 170K cases, selecting those with the same case name as positive samples and others as negative. To ensure comparability with other baselines, we use 5 positive and 5 negative cases per query.

\paragraph{KELLER.}
We implement KELLER based on the code from the official repository\footnote{\protect\url{https://github.com/ChenlongDeng/KELLER}}.
To prepare the retrieval pool, we first process the same 1.2M cases used in implementing SAILER. 
To further separate cases into subfactual levels, we use GPT-4o. 
Each subfact is labeled with its criminal type using either their subheadings or regular expressions. This results in 1.1M cases.

For training, we use the same query cases from SearchLM. Following the same process, we separate and label each subfact, resulting in 143K cases. 
We prepare ground truth document cases by matching cases that include the same subfacts criminal type as the query case. Among the matching cases, we use the top 10 document cases based on BM25 scores. 
This results in total of 820k (query case, document case) pairs. 
The model is trained on the resulting dataset for 1 epoch under the same condition with the original Keller paper (batch size: 128, learning rate: 1e-5, optimizer: AdamW).

\section{Licenses and intended use}

Korean legal cases are not protected by the Korean Copyright Act. mT5 \citep{xue-etal-2021-mt5}, the base model of \textsc{LegalSearchLM}, is disclosed with Apache 2.0 license that permits free academic use.

Korean legal cases are fully anonymized when disclosed by the Korean court. However, we do not censor potentially offensive content, including descriptions about violent and sexual crimes, as they constitute the core content of legal cases.

\keepXColumns
\renewcommand{\arraystretch}{1.3} 
\onecolumn
\scriptsize

\begin{tabularx}{\textwidth}{X X}
\caption{List of query case types for Sexual Crime}\label{tab:korean_sexual_crime} \\
\hline
\textbf{Law} & \textbf{Crime} \\
\hline

형법 제297조 & 강간 \\
형법 제297조의2 & 유사강간 \\
형법 제298조 & 강제추행 \\
형법 제299조, 제297조 & 준강간 \\
형법 제299조, 제297조의2 & 준유사강간 \\
형법 제299조, 제298조 & 준강제추행 \\
형법 제300조, 제297조 & 강간미수 \\
형법 제300조, 제297조의2 & 유사강간미수 \\
형법 제300조, 제298조 & 강제추행미수 \\
형법 제300조, 제297조, 제299조 & 준강간미수 \\
형법 제300조, 제297조의2, 제299조 & 준유사강간미수 \\
형법 제300조, 제298조, 제299조 & 준강제추행미수 \\
형법 제301조, 제297조 & 강간상해치상 \\
형법 제301조, 제297조의2 & 유사강간상해치상 \\
형법 제301조, 제298조 & 강제추행상해치상 \\
형법 제301조, 제297조, 제299조 & 준강간상해치상 \\
형법 제301조, 제297조의2, 제299조 & 준유사강간상해치상 \\
형법 제301조, 제298조, 제299조 & 준강제추행상해치상 \\
형법 제301조, 제300조, 제297조 & 강간미수상해치상 \\
형법 제301조, 제300조, 제297조의2 & 유사강간미수상해치상 \\
형법 제301조, 제300조, 제298조 & 강제추행미수상해치상 \\
형법 제301조, 제300조, 제297조, 제299조 & 준강간미수상해치상 \\
형법 제301조, 제300조, 제297조의2, 제299조 & 준유사강간미수상해치상 \\
형법 제301조, 제300조, 제298조, 제299조 & 준강제추행미수상해치상 \\
형법 제301조의2, 제297조 & 강간살인치사 \\
형법 제301조의2, 제297조의2 & 유사강간살인치사 \\
형법 제301조의2, 제298조 & 강제추행살인치사 \\
형법 제301조의2, 제297조, 제299조 & 준강간살인치사 \\
형법 제301조의2, 제297조의2, 제299조 & 준유사강간살인치사 \\
형법 제301조의2, 제298조, 제299조 & 준강제추행살인치사 \\
형법 제301조의2, 제300조, 제297조 & 강간미수살인치사 \\
형법 제301조의2, 제300조, 제297조의2 & 유사강간미수살인치사 \\
형법 제301조의2, 제300조, 제298조 & 강제추행미수살인치사 \\
형법 제301조의2, 제300조, 제297조, 제299조 & 준강간미수살인치사 \\
형법 제301조의2, 제300조, 제297조의2, 제299조 & 준유사강간미수살인치사 \\
형법 제301조의2, 제300조, 제298조, 제299조 & 준강제추행미수살인치사 \\
형법 제302조 & 미성년자간음 / 심신미약자간음 / 미성년자추행 / 심신미약자추행 \\
형법 제303조 제1항 & 피보호자간음 / 피감독자간음 \\
형법 제305조, 제297조 & 미성년자의제강간 \\
형법 제305조, 제297조의2 & 미성년자의제유사강간 \\
형법 제305조, 제298조 & 미성년자의제강제추행 \\
형법 제305조, 제301조, 제297조 & 미성년자의제강간상해치상 \\
형법 제305조, 제301조, 제297조의2 & 미성년자의제유사강간상해치상 \\
형법 제305조, 제301조, 제298조 & 미성년자의제강제추행상해치상 \\
형법 제305조, 제301조의2, 제297조 & 미성년자의제강간살인치사 \\
형법 제305조, 제301조의2, 제297조의2 & 미성년자의제유사강간살인치사 \\
형법 제305조, 제301조의2, 제298조 & 미성년자의제강제추행살인치사 \\
형법 제305조의3, 제297조 & 강간예비음모 \\
형법 제305조의3, 제297조의2 & 유사강간예비음모 \\
형법 제305조의3, 제297조, 제299조 & 준강간예비음모 \\
형법 제305조의3, 제301조, 제297조 & 강간상해예비음모 \\
형법 제305조의3, 제301조, 제298조 & 강제추행상해예비음모 \\
형법 제305조의3, 제301조, 제297조, 제299조 & 준강간상해예비음모 \\
형법 제305조의3, 제301조, 제297조의2, 제299조 & 준유사강간상해예비음모 \\
형법 제305조의3, 제301조, 제298조, 제299조 & 준강제추행상해예비음모 \\
형법 제305조의3, 제305조, 제297조 & 미성년자의제강간예비음모 \\
형법 제305조의3, 제305조, 제297조의2 & 미성년자의제유사강간예비음모 \\
형법 제305조의3, 제305조, 제298조 & 미성년자의제강제추행예비음모 \\
형법 제305조의3, 제305조, 제301조, 제297조 & 미성년자의제강간상해치상예비음모 \\
형법 제305조의3, 제305조, 제301조, 제297조의2 & 미성년자의제유사강간상해치상예비음모 \\
형법 제305조의3, 제305조, 제301조, 제298조 & 미성년자의제강제추행상해치상예비음모 \\
형법 제305조의3, 제305조, 제301조의2, 제297조 & 미성년자의제강간살인치사예비음모 \\
형법 제305조의3, 제305조, 제301조의2, 제297조의2 & 미성년자의제유사강간살인치사예비음모 \\
형법 제305조의3, 제305조, 제301조의2, 제298조 & 미성년자의제강제추행살인치사예비음모 \\

성폭력범죄의처벌등에관한특례법 제3조 제1항, 형법 제319조 제1항, 제297조 & 주거침입강간 \\
성폭력범죄의처벌등에관한특례법 제3조 제1항, 형법 제319조 제1항, 제297조의2 & 주거침입유사강간 \\
성폭력범죄의처벌등에관한특례법 제3조 제1항, 형법 제319조 제1항, 제298조 & 주거침입강제추행 \\
성폭력범죄의처벌등에관한특례법 제3조 제1항, 형법 제319조 제1항, 제299조, 제297조 & 주거침입준강간 \\
성폭력범죄의처벌등에관한특례법 제3조 제1항, 형법 제319조 제1항, 제299조, 제297조의2 & 주거침입준유사강간 \\
성폭력범죄의처벌등에관한특례법 제3조 제1항, 형법 제319조 제1항, 제299조, 제298조 & 주거침입준강제추행 \\
성폭력범죄의처벌등에관한특례법 제3조 제1항, 형법 제330조, 제297조 & 야간주거침입절도강간 \\
성폭력범죄의처벌등에관한특례법 제3조 제1항, 형법 제330조, 제297조의2 & 야간주거침입절도유사강간 \\
성폭력범죄의처벌등에관한특례법 제3조 제1항, 형법 제330조, 제298조 & 야간주거침입절도강제추행 \\
성폭력범죄의처벌등에관한특례법 제3조 제1항, 형법 제330조, 제299조, 제297조 & 야간주거침입절도준강간 \\
성폭력범죄의처벌등에관한특례법 제3조 제1항, 형법 제330조, 제299조, 제297조의2 & 야간주거침입절도준유사강간 \\
성폭력범죄의처벌등에관한특례법 제3조 제1항, 형법 제330조, 제299조, 제298조 & 야간주거침입절도준강제추행 \\
성폭력범죄의처벌등에관한특례법 제3조 제1항, 형법 제331조, 제297조 & 특수절도강간 \\
성폭력범죄의처벌등에관한특례법 제3조 제1항, 형법 제331조, 제297조의2 & 특수절도유사강간 \\
성폭력범죄의처벌등에관한특례법 제3조 제1항, 형법 제331조, 제298조 & 특수절도강제추행 \\
성폭력범죄의처벌등에관한특례법 제3조 제1항, 형법 제331조, 제299조, 제297조 & 특수절도준강간 \\
성폭력범죄의처벌등에관한특례법 제3조 제1항, 형법 제331조, 제299조, 제297조의2 & 특수절도준유사강간 \\
성폭력범죄의처벌등에관한특례법 제3조 제1항, 형법 제331조, 제299조, 제298조 & 특수절도준강제추행 \\
성폭력범죄의처벌등에관한특례법 제3조 제1항, 형법 제342조, 제330조, 제297조 & 야간주거침입절도미수강간 \\
성폭력범죄의처벌등에관한특례법 제3조 제1항, 형법 제342조, 제330조, 제297조의2 & 야간주거침입절도미수유사강간 \\
성폭력범죄의처벌등에관한특례법 제3조 제1항, 형법 제342조, 제330조, 제298조 & 야간주거침입절도미수강제추행 \\
성폭력범죄의처벌등에관한특례법 제3조 제1항, 형법 제342조, 제330조, 제299조, 제297조 & 야간주거침입절도미수준강간 \\
성폭력범죄의처벌등에관한특례법 제3조 제1항, 형법 제342조, 제330조, 제299조, 제297조의2 & 야간주거침입절도미수준유사강간 \\
성폭력범죄의처벌등에관한특례법 제3조 제1항, 형법 제342조, 제330조, 제299조, 제298조 & 야간주거침입절도미수준강제추행 \\
성폭력범죄의처벌등에관한특례법 제3조 제1항, 형법 제342조, 제331조, 제297조 & 특수절도미수강간 \\
성폭력범죄의처벌등에관한특례법 제3조 제1항, 형법 제342조, 제331조, 제297조의2 & 특수절도미수유사강간 \\
성폭력범죄의처벌등에관한특례법 제3조 제1항, 형법 제342조, 제331조, 제298조 & 특수절도미수강제추행 \\
성폭력범죄의처벌등에관한특례법 제3조 제1항, 형법 제342조, 제331조, 제299조, 제297조 & 특수절도미수준강간 \\
성폭력범죄의처벌등에관한특례법 제3조 제1항, 형법 제342조, 제331조, 제299조, 제297조의2 & 특수절도미수준유사강간 \\
성폭력범죄의처벌등에관한특례법 제3조 제1항, 형법 제342조, 제331조, 제299조, 제298조 & 특수절도미수준강제추행 \\

성폭력범죄의처벌등에관한특례법 제3조 제2항, 형법 제319조 제1항, 제297조 & 특수강도강간 \\
성폭력범죄의처벌등에관한특례법 제3조 제2항, 형법 제334조, 제297조의2 & 특수강도유사강간 \\
성폭력범죄의처벌등에관한특례법 제3조 제2항, 형법 제334조, 제298조 & 특수강도강제추행 \\
성폭력범죄의처벌등에관한특례법 제3조 제2항, 형법 제334조, 제299조, 제297조 & 특수강도준강간 \\
성폭력범죄의처벌등에관한특례법 제3조 제2항, 형법 제334조, 제299조, 제298조 & 특수강도준강제추행 \\
성폭력범죄의처벌등에관한특례법 제3조 제2항, 형법 제342조, 제334조, 제297조 & 특수강도미수강간 \\
성폭력범죄의처벌등에관한특례법 제3조 제2항, 형법 제342조, 제334조, 제297조의2 & 특수강도미수유사강간 \\
성폭력범죄의처벌등에관한특례법 제3조 제2항, 형법 제342조, 제334조, 제298조 & 특수강도미수강제추행 \\
성폭력범죄의처벌등에관한특례법 제3조 제2항, 형법 제342조, 제334조, 제299조, 제297조 & 특수강도미수준강간 \\
성폭력범죄의처벌등에관한특례법 제3조 제2항, 형법 제342조, 제334조, 제299조, 제297조의2 & 특수강도미수준유사강간 \\
성폭력범죄의처벌등에관한특례법 제3조 제2항, 형법 제342조, 제334조, 제299조, 제298조 & 특수강도미수준강제추행 \\

성폭력범죄의처벌등에관한특례법 제4조 제1항 & 성폭력범죄의처벌등에관한특례법위반(특수강간) \\
성폭력범죄의처벌등에관한특례법 제4조 제2항 & 성폭력범죄의처벌등에관한특례법위반(특수강제추행) \\
성폭력범죄의처벌등에관한특례법 제4조 제3항 & 성폭력범죄의처벌등에관한특례법위반(특수준강간) \\
성폭력범죄의처벌등에관한특례법 제4조 제3항 & 성폭력범죄의처벌등에관한특례법위반(특수준강제추행) \\
성폭력범죄의처벌등에관한특례법 제5조 제1항 & 성폭력범죄의처벌등에관한특례법위반(친족관계에의한강간) \\
성폭력범죄의처벌등에관한특례법 제5조 제2항 & 성폭력범죄의처벌등에관한특례법위반(친족관계에의한강제추행) \\
성폭력범죄의처벌등에관한특례법 제5조 제3항 & 성폭력범죄의처벌등에관한특례법위반(친족관계에의한준강간) \\
성폭력범죄의처벌등에관한특례법 제5조 제3항 & 성폭력범죄의처벌등에관한특례법위반(친족관계에의한준강제추행) \\
성폭력범죄의처벌등에관한특례법 제6조 제1항 & 성폭력범죄의처벌등에관한특례법위반(장애인강간) \\
성폭력범죄의처벌등에관한특례법 제6조 제2항 & 성폭력범죄의처벌등에관한특례법위반(장애인유사성행위) \\
성폭력범죄의처벌등에관한특례법 제6조 제3항 & 성폭력범죄의처벌등에관한특례법위반(장애인강제추행) \\
성폭력범죄의처벌등에관한특례법 제6조 제4항 & 성폭력범죄의처벌등에관한특례법위반(장애인준강간) \\
성폭력범죄의처벌등에관한특례법 제6조 제4항 & 성폭력범죄의처벌등에관한특례법위반(장애인준유사성행위) \\
성폭력범죄의처벌등에관한특례법 제6조 제4항 & 성폭력범죄의처벌등에관한특례법위반(장애인준강제추행) \\
성폭력범죄의처벌등에관한특례법 제6조 제5항 & 성폭력범죄의처벌등에관한특례법위반(장애인위계등간음) \\
성폭력범죄의처벌등에관한특례법 제6조 제6항 & 성폭력범죄의처벌등에관한특례법위반(장애인위계등추행) \\
성폭력범죄의처벌등에관한특례법 제6조 제7항 & 성폭력범죄의처벌등에관한특례법위반(장애인피보호자강간등) \\
성폭력범죄의처벌등에관한특례법 제7조 제1항 & 성폭력범죄의처벌등에관한특례법위반(13세미만미성년자강간) \\
성폭력범죄의처벌등에관한특례법 제7조 제2항 & 성폭력범죄의처벌등에관한특례법위반(13세미만미성년자유사성행위) \\
성폭력범죄의처벌등에관한특례법 제7조 제3항 & 성폭력범죄의처벌등에관한특례법위반(13세미만미성년자강제추행) \\
성폭력범죄의처벌등에관한특례법 제7조 제4항 & 성폭력범죄의처벌등에관한특례법위반(13세미만미성년자준강간) \\
성폭력범죄의처벌등에관한특례법 제7조 제4항 & 성폭력범죄의처벌등에관한특례법위반(13세미만미성년자준유사성행위) \\
성폭력범죄의처벌등에관한특례법 제7조 제4항 & 성폭력범죄의처벌등에관한특례법위반(13세미만미성년자준강제추행) \\
성폭력범죄의처벌등에관한특례법 제7조 제5항 & 성폭력범죄의처벌등에관한특례법위반(13세미만미성년자위계등간음) \\
성폭력범죄의처벌등에관한특례법 제7조 제5항 & 성폭력범죄의처벌등에관한특례법위반(13세미만미성년자위계등유사성행위) \\
성폭력범죄의처벌등에관한특례법 제7조 제5항 & 성폭력범죄의처벌등에관한특례법위반(13세미만미성년자위계등추행) \\

성폭력범죄의처벌등에관한특례법 제8조 &	성폭력범죄의처벌등에관한특례법위반(강간등상해) / 성폭력범죄의처벌등에관한특례법위반(강간등치상) \\
성폭력범죄의처벌등에관한특례법 제9조	& 성폭력범죄의처벌등에관한특례법위반(강간등살인) / 성폭력범죄의처벌등에관한특례법위반(강간등치사) \\

성폭력범죄의처벌등에관한특례법 제10조 & 성폭력범죄의처벌등에관한특례법위반(업무상위력등에의한추행) \\
성폭력범죄의처벌등에관한특례법 제11조 & 성폭력범죄의처벌등에관한특례법위반(공중밀집장소에서의추행) \\
성폭력범죄의처벌등에관한특례법 제12조 & 성폭력범죄의처벌등에관한특례법위반(성적목적다중이용장소침입) \\
성폭력범죄의처벌등에관한특례법 제13조 & 성폭력범죄의처벌등에관한특례법위반(통신매체이용음란) \\
성폭력범죄의처벌등에관한특례법 제14조 제1항 & 성폭력범죄의처벌등에관한특례법위반(카메라등이용촬영) \\
성폭력범죄의처벌등에관한특례법 제14조 제2항 & 성폭력범죄의처벌등에관한특례법위반(카메라등이용촬영물반포등) \\
성폭력범죄의처벌등에관한특례법 제14조 제3항 & 성폭력범죄의처벌등에관한특례법위반(영리목적카메라등이용촬영물반포등) \\
성폭력범죄의처벌등에관한특례법 제14조 제4항 & 성폭력범죄의처벌등에관한특례법위반(카메라등이용촬영물소지등) \\
성폭력범죄의처벌등에관한특례법 제14조 제5항 & 성폭력범죄의처벌등에관한특례법위반(상습카메라등이용촬영·반포등) \\
성폭력범죄의처벌등에관한특례법 제14조의2 제1항 & 성폭력범죄의처벌등에관한특례법위반(허위영상물편집등) \\
성폭력범죄의처벌등에관한특례법 제14조의2 제2항 & 성폭력범죄의처벌등에관한특례법위반(허위영상물반포등) \\
성폭력범죄의처벌등에관한특례법 제14조의2 제3항 & 성폭력범죄의처벌등에관한특례법위반(영리목적허위영상물반포등) \\
성폭력범죄의처벌등에관한특례법 제14조의3 제1항 & 성폭력범죄의처벌등에관한특례법위반(촬영물등이용협박) \\
성폭력범죄의처벌등에관한특례법 제14조의3 제2항 & 성폭력범죄의처벌등에관한특례법위반(촬영물등이용강요) \\

성폭력범죄의처벌등에관한특례법 제14조의3 제3항, 제14조의3 제1항, 제14조 제1항 & 성폭력범죄의처벌등에관한특례법위반(상습촬영물등이용협박) \\
성폭력범죄의처벌등에관한특례법 제50조 & 성폭력범죄의처벌등에관한특례법위반(비밀준수등) \\
성매매알선등행위의처벌에관한법률 제19조 & 성매매알선등행위의처벌에관한법률위반(성매매알선등) \\
성매매알선등행위의처벌에관한법률 제20조 & 성매매알선등행위의처벌에관한법률위반(성매매광고) \\ 
성매매알선등행위의처벌에관한법률 제21조 제1항 & 성매매알선등행위의처벌에관한법률위반(성매매) \\
아동·청소년의성보호에관한법률 제7조 제1항	& 아동·청소년의성보호에관한법률위반(강간) \\
아동·청소년의성보호에관한법률 제7조 제2항	& 아동·청소년의성보호에관한법률위반(유사성행위) \\
아동·청소년의성보호에관한법률 제7조 제3항	& 아동·청소년의성보호에관한법률위반(강제추행) \\
아동·청소년의성보호에관한법률 제7조 제4항	& 아동·청소년의성보호에관한법률위반(준강간) \\
아동·청소년의성보호에관한법률 제7조 제4항	& 아동·청소년의성보호에관한법률위반(준유사성행위) \\
아동·청소년의성보호에관한법률 제7조 제4항	& 아동·청소년의성보호에관한법률위반(준강제추행) \\ 
아동·청소년의성보호에관한법률 제7조 제2항	& 아동·청소년의성보호에관한법률위반(위계등간음) \\
아동·청소년의성보호에관한법률 제7조 제2항	& 아동·청소년의성보호에관한법률위반(위계등유사성행위) \\
아동·청소년의성보호에관한법률 제7조 제2항	& 아동·청소년의성보호에관한법률위반(위계등추행) \\
아동·청소년의성보호에관한법률 제7조의2, 제7조 제1항	& 아동·청소년의성보호에관한법률위반(강간예비음모) \\ 
아동·청소년의성보호에관한법률 제7조의2, 제7조 제2항	& 아동·청소년의성보호에관한법률위반(유사성행위예비음모) \\
아동·청소년의성보호에관한법률 제7조의2, 제7조 제3항	& 아동·청소년의성보호에관한법률위반(강제추행예비음모) \\
아동·청소년의성보호에관한법률 제7조의2, 제7조 제4항	& 아동·청소년의성보호에관한법률위반(준강간예비음모) \\
아동·청소년의성보호에관한법률 제7조의2, 제7조 제4항	& 아동·청소년의성보호에관한법률위반(준유사성행위예비음모) \\
아동·청소년의성보호에관한법률 제7조의2, 제7조 제4항	& 아동·청소년의성보호에관한법률위반(준강제추행예비음모) \\
아동·청소년의성보호에관한법률 제7조의2, 제7조 제2항	& 아동·청소년의성보호에관한법률위반(위계등간음예비음모) \\
아동·청소년의성보호에관한법률 제7조의2, 제7조 제2항	& 아동·청소년의성보호에관한법률위반(위계등유사성행위예비음모) \\
아동·청소년의성보호에관한법률 제7조의2, 제7조 제2항	아동·청소년의성보호에관한법률위반(위계등추행예비음모) \\

아동·청소년의성보호에관한법률 제8조 제1항 & 아동·청소년의성보호에관한법률위반(장애인간음) \\
아동·청소년의성보호에관한법률 제8조 제2항 & 아동·청소년의성보호에관한법률위반(장애인추행) \\
아동·청소년의성보호에관한법률 제8조의2 제1항 & 아동·청소년의성보호에관한법률위반(16세미만아동·청소년간음) \\
아동·청소년의성보호에관한법률 제8조의2 제2항 & 아동·청소년의성보호에관한법률위반(16세미만아동·청소년추행) \\
아동·청소년의성보호에관한법률 제9조 & 아동·청소년의성보호에관한법률위반(강간등상해) / 아동·청소년의성보호에관한법률위반(강간등치상) \\
아동·청소년의성보호에관한법률 제10조 & 아동·청소년의성보호에관한법률위반(강간등살인) / 아동·청소년의성보호에관한법률위반(강간등치사) \\
아동·청소년의성보호에관한법률 제11조 제1항 & 아동·청소년의성보호에관한법률위반(성착취물제작등) \\
아동·청소년의성보호에관한법률 제11조 제2항 & 아동·청소년의성보호에관한법률위반(영리목적성착취물판매등) \\
아동·청소년의성보호에관한법률 제11조 제3항 & 아동·청소년의성보호에관한법률위반(성착취물배포등) \\
아동·청소년의성보호에관한법률 제11조 제5항 & 아동·청소년의성보호에관한법률위반(성착취물소지등) \\
아동·청소년의성보호에관한법률 제11조 제7항 & 아동·청소년의성보호에관한법률위반(상습성착취물제작·배포등) \\
아동·청소년의성보호에관한법률 제13조 제1항 & 아동·청소년의성보호에관한법률위반(성매수등) \\
아동·청소년의성보호에관한법률 제15조의2 & 아동·청소년의성보호에관한법률위반(성착취목적대화등) \\

스토킹범죄의 처벌 등에 관한 법률 제18조 제1항	& 스토킹범죄의처벌등에관한법률위반 \\
스토킹범죄의 처벌 등에 관한 법률 제20조 & 스토킹범죄의처벌등에관한법률위반 \\

\hline
\end{tabularx}
\renewcommand{\arraystretch}{1.3} 

\scriptsize
\begin{longtable}{p{0.47\textwidth} p{0.47\textwidth}}
\caption{List of query case types for Fraud}\label{tab:korean_fraud}\\
\hline
\textbf{Law} & \textbf{Crime}\\
\hline
\endfirsthead
\hline
\textbf{Law} & \textbf{Crime}\\
\hline
\endhead
\hline
\endfoot
형법 제347조 & 사기 \\
형법 제347조, 제351조 & 상습사기 \\
형법 제347조, 제352조 & 사기미수 \\
형법 제347조, 제351조, 제352조 & 상습사기미수 \\
형법 제347조의2 & 컴퓨터등사용사기 \\
형법 제347조의2, 제351조 & 상습컴퓨터등사용사기 \\
형법 제347조의2, 제352조 & 컴퓨터등사용사기미수 \\
형법 제347조의2, 제351조, 제352조 & 상습컴퓨터등사용사기미수 \\
형법 제348조 & 준사기 \\
형법 제348조, 제351조 & 상습준사기 \\
형법 제348조, 제352조 & 준사기미수 \\
형법 제348조, 제351조, 제352조 & 상습준사기미수 \\
형법 제348조의2 & 편의시설부정이용 \\
형법 제348조의2, 제351조 & 상습편의시설부정이용 \\
형법 제348조의2, 제352조 & 편의시설부정이용미수 \\
형법 제348조의2, 제351조, 제352조 & 상습편의시설부정이용미수 \\
형법 제350조 & 공갈 \\
형법 제350조, 제351조 & 상습공갈 \\
형법 제350조, 제352조 & 공갈미수 \\
형법 제350조, 제351조, 제352조 & 상습공갈미수 \\
형법 제350조의2 & 특수공갈 \\
형법 제350조의2, 제351조 & 상습특수공갈 \\
형법 제350조의2, 제352조 & 특수공갈미수 \\
형법 제350조의2, 제351조, 제352조 & 상습특수공갈미수 \\
폭력행위등처벌에관한법률 제2조 제2항, 형법 제350조 & 폭력행위등처벌에관한법률위반(공갈) 
\\
폭력행위등처벌에관한법률 제2조 제3항, 형법 제350조 & 폭력행위등처벌에관한법률위반(공갈재범) \\
특정경제범죄가중처벌등에관한법률 제3조, 형법 제347조, 제347조의2, 제351조 & 특정경제범죄가중처벌등에관한법률위반(사기) \\
특정경제범죄가중처벌등에관한법률 제3조, 형법 제350조, 제350조의2, 제351조 & 특정경제범죄가중처벌등에관한법률위반(공갈) \\
여신전문금융업법 제70조 & 여신전문금융업법위반 \\
\hline
\end{longtable}
\renewcommand{\arraystretch}{1.3} 

\onecolumn       
\scriptsize
\begin{longtable}{p{0.47\textwidth} p{0.47\textwidth}}
\caption{List of query case types for Theft and Robbery}\label{tab:korean_theft_robbery}\\
\hline
\textbf{Law} & \textbf{Crime}\\
\hline
\endfirsthead
\hline
\textbf{Law} & \textbf{Crime}\\
\hline
\endhead
\hline
\endfoot
형법 제329조 & 절도 \\
형법 제329조, 제342조 & 절도미수 \\
형법 제329조, 제332조 & 상습절도 \\
형법 제329조, 제332조, 제342조 & 상습절도미수 \\
형법 제330조 & 야간주거침입절도 \\
형법 제330조, 제342조 & 야간주거침입절도미수 \\
형법 제330조, 제332조 & 상습야간주거침입절도 \\
형법 제330조, 제332조, 제342조 & 상습야간주거침입절도미수 \\
형법 제330조 & 야간건조물침입절도 \\
형법 제330조, 제342조 & 야간건조물침입절도미수 \\
형법 제330조, 제332조 & 상습야간건조물침입절도 \\
형법 제330조, 제332조, 제342조 & 상습야간건조물침입절도미수 \\
형법 제330조 & 야간선박침입절도 \\
형법 제330조, 제342조 & 야간선박침입절도미수 \\
형법 제330조, 제332조 & 상습야간선박침입절도 \\
형법 제330조, 제332조, 제342조 & 상습야간선박침입절도미수 \\
형법 제330조 & 야간항공기침입절도 \\
형법 제330조, 제342조 & 야간항공기침입절도미수 \\
형법 제330조, 제332조 & 상습야간항공기침입절도 \\
형법 제330조, 제332조, 제342조 & 상습야간항공기침입절도미수 \\
형법 제330조 & 야간방실침입절도 \\
형법 제330조, 제342조 & 야간방실침입절도미수 \\
형법 제330조, 제332조 & 상습야간방실침입절도 \\
형법 제330조, 제332조, 제342조 & 상습야간방실침입절도미수 \\
형법 제331조 & 특수절도 \\
형법 제331조, 제342조 & 특수절도미수 \\
형법 제331조, 제332조 & 상습특수절도 \\
형법 제331조, 제332조, 제342조 & 상습특수절도미수 \\
형법 제333조 & 강도 \\
형법 제333조, 제342조 & 강도미수 \\
형법 제333조, 제341조 & 상습강도 \\
형법 제333조, 제341조, 제342조 & 상습강도미수 \\
형법 제334조 & 특수강도 \\
형법 제334조, 제342조 & 특수강도미수 \\
형법 제334조, 제341조 & 상습특수강도 \\
형법 제334조, 제341조, 제342조 & 상습특수강도미수 \\
형법 제335조 & 준강도 \\
형법 제335조, 제342조 & 준강도미수 \\
형법 제335조 & 준특수강도 \\
형법 제335조, 제342조 & 준특수강도미수 \\
형법 제337조 & 강도상해 \\
형법 제337조, 제342조 & 강도상해미수 \\
형법 제337조 & 강도치상 \\
형법 제337조, 제342조 & 강도치상미수 \\
형법 제338조 & 강도살인 \\
형법 제338조, 제342조 & 강도살인미수 \\
형법 제338조 & 강도치사 \\
형법 제338조, 제342조 & 강도치사미수 \\
형법 제339조 & 강도강간 \\
형법 제339조, 제342조 & 강도강간미수 \\
형법 제343조 & 강도예비 \\
형법 제343조 & 강도음모 \\
\hline
\end{longtable}
\renewcommand{\arraystretch}{1.3} 

\onecolumn       
\scriptsize
\begin{longtable}{p{0.47\textwidth} p{0.47\textwidth}}
\caption{List of query case types for Embezzlement}\label{tab:korean_embezzlement}\\
\hline
\textbf{Law} & \textbf{Crime}\\
\hline
\endfirsthead
\hline
\textbf{Law} & \textbf{Crime}\\
\hline
\endhead
\hline
\endfoot
형법 제355조 제1항 & 횡령 \\
형법 제355조 제1항, 제359조 & 횡령미수 \\
형법 제355조 제2항 & 배임 \\
형법 제355조 제2항, 제359조 & 배임미수 \\
형법 제356조 & 업무상횡령 \\
형법 제356조, 제359조 & 업무상횡령미수 \\
형법 제356조 & 업무상배임 \\
형법 제356조, 제359조 & 업무상배임미수 \\
형법 제357조 제1항 & 배임수재 \\
형법 제357조 제1항, 제359조 & 배임수재미수 \\
형법 제357조 제2항 & 배임증재 \\
형법 제357조 제2항, 제359조 & 배임증재미수 \\
형법 제360조 제1항 & 점유이탈물횡령 \\
형법 제360조 제1항, 제359조 & 점유이탈물횡령미수 \\
특정경제범죄가중처벌등에관한법률 제3조, 형법 제355조 제1항, 제356조 & 특정경제범죄가중처벌등에관한법률위반(횡령) \\
특정경제범죄가중처벌등에관한법률 제3조, 형법 제355조 제2항, 제356조 & 특정경제범죄가중처벌등에관한법률위반(배임) \\
\hline
\end{longtable}
\renewcommand{\arraystretch}{1.3} 

\scriptsize
\begin{longtable}{p{0.47\textwidth} p{0.47\textwidth}}
\caption{List of query case types for Destruction}\label{tab:korean_destruction}\\
\hline
\textbf{Law} & \textbf{Crime}\\
\hline
\endfirsthead
\hline
\textbf{Law} & \textbf{Crime}\\
\hline
\endhead
\hline
\endfoot
형법 제366조 & 재물손괴 \\
형법 제369조, 제366조 & 특수재물손괴 \\
형법 제369조, 제371조 & 특수재물손괴미수 \\
형법 제371조, 제366조 & 재물손괴미수 \\
형법 제371조, 제369조, 제366조 & 특수재물손괴미수 \\
폭력행위등처벌에관한법률 제2조 제2항, 형법 제366조 & 폭력행위등처벌에관한법률위반(재물손괴등) \\
폭력행위등처벌에관한법률 제2조 제3항, 형법 제366조 & 폭력행위등처벌에관한법률위반(재물손괴등재범) \\
\hline
\end{longtable}
\renewcommand{\arraystretch}{1.3} 

\scriptsize
\begin{longtable}{p{0.47\textwidth} p{0.47\textwidth}}
\caption{List of query case types for Murder}\label{tab:korean_murder}\\
\hline
\textbf{Law} & \textbf{Crime}\\
\hline
\endfirsthead
\hline
\textbf{Law} & \textbf{Crime}\\
\hline
\endhead
\hline
\endfoot
형법 제250조 제1항	& 살인 \\
형법 제254조, 제250조 제1항	& 살인미수 \\
\hline
\end{longtable}
\renewcommand{\arraystretch}{1.3} 

\scriptsize
\begin{longtable}{p{0.47\textwidth} p{0.47\textwidth}}
\caption{List of query case types for Injury and Violence}\label{tab:korean_violence}\\
\hline
\textbf{Law} & \textbf{Crime}\\
\hline
\endfirsthead
\hline
\textbf{Law} & \textbf{Crime}\\
\hline
\endhead
\hline
\endfoot
형법 제257조 제1항 & 상해 \\
형법 제257조 제1항, 제264조 & 상습상해 \\
형법 제257조 제3항, 제1항 & 상해미수 \\
형법 제258조 제1항/제258조 제2항 & 중상해 \\
형법 제258조, 제264조 & 상습중상해 \\
형법 제258조의2 제1항, 제257조 제1항 & 특수상해 \\
형법 제258조의2, 제264조 & 상습특수상해 \\
형법 제258조의2 제3항, 제1항, 제257조 제1항 & 특수상해미수 \\
형법 제258조의2, 제264조 & 상습특수상해미수 \\
형법 제259조 제1항 & 상해치사 \\
형법 제260조 제1항 & 폭행 \\
형법 제260조 제1항, 제264조 & 상습폭행 \\
형법 제261조, 제260조 제1항 & 특수폭행 \\
형법 제261조, 제264조 & 상습특수폭행 \\
형법 제262조, 제260조 제1항, (제259조, 제257조 제1항) & 폭행치사 \\
형법 제262조, 제260조 제1항, (제259조, 제257조 제1항) & 폭행치상 \\
형법 제262조, 제261조, (제259조, 제257조 제1항) & 특수폭행치사 \\
형법 제262조, 제261조, (제259조, 제257조 제1항) & 특수폭행치상 \\
폭력행위등처벌에관한법률 제2조 제2항 & 폭력행위등처벌에관한법률위반(폭행) \\
폭력행위등처벌에관한법률 제2조 제2항 & 폭력행위등처벌에관한법률위반(존속폭행) \\
폭력행위등처벌에관한법률 제2조 제2항 & 폭력행위등처벌에관한법률위반(상해) \\
폭력행위등처벌에관한법률 제2조 제2항 & 폭력행위등처벌에관한법률위반(존속상해) \\
폭력행위등처벌에관한법률 제2조 제3항 & 폭력행위등처벌에관한법률위반(폭행재범) \\
폭력행위등처벌에관한법률 제2조 제3항 & 폭력행위등처벌에관한법률위반(존속폭행재범) \\
폭력행위등처벌에관한법률 제2조 제3항 & 폭력행위등처벌에관한법률위반(상해재범) \\
폭력행위등처벌에관한법률 제2조 제3항 & 폭력행위등처벌에관한법률위반(존속상해재범) \\
폭력행위등처벌에관한법률 제4조 제1항 & 폭력행위등처벌에관한법률위반(단체등의구성·활동) \\
형법 제262조, 제260조 제2항 & 존속폭행치상 \\
형법 제262조, 제260조 제2항 & 존속폭행치사 \\
형법 제262조, 제260조 제2항 & 존속상해치상 \\
형법 제262조, 제260조 제2항 & 존속상해치사 \\
형법 제260조 제2항, 제264조 & 상습존속폭행 \\
형법 제260조 제2항, 제264조 & 상습존속상해 \\
형법 제261조, 제260조 제2항 & 특수존속폭행 \\
형법 제261조, 제260조 제2항 & 특수존속상해 \\
형법 제260조 제2항 & 존속폭행 \\
형법 제260조 제2항 & 존속상해 \\
형법 제258조 제3항 & 존속중상해 \\

\hline
\end{longtable}
\renewcommand{\arraystretch}{1.3} 

\scriptsize
\begin{longtable}{p{0.47\textwidth} p{0.47\textwidth}}
\caption{List of query case types for Negligent homicide and injury}\label{tab:korean_negligent}\\
\hline
\textbf{Law} & \textbf{Crime}\\
\hline
\endfirsthead
\hline
\textbf{Law} & \textbf{Crime}\\
\hline
\endhead
\hline
\endfoot
형법 제266조 & 과실치상 \\
형법 제267조 & 과실치사 \\
형법 제268조 & 업무상과실치사 \\
형법 제268조 & 중과실치사 \\
형법 제268조 & 업무상과실치상 \\
형법 제268조 & 중과실치상 \\
\hline
\end{longtable}
\renewcommand{\arraystretch}{1.3} 

\scriptsize
\begin{longtable}{p{0.47\textwidth} p{0.47\textwidth}}
\caption{List of query case types for Arrest and Detention}\label{tab:korean_arrest}\\
\hline
\textbf{Law} & \textbf{Crime}\\
\hline
\endfirsthead
\hline
\textbf{Law} & \textbf{Crime}\\
\hline
\endhead
\hline
\endfoot
형법 제276조 제1항 & 체포 \\
형법 제276조 제1항 & 감금 \\
형법 제277조 제1항 & 중체포 \\
형법 제277조 제1항 & 중감금 \\
형법 제276조 제1항, 제280조 & 체포미수 \\
형법 제276조 제1항, 제280조 & 감금미수 \\
형법 제277조 제1항, 제280조 & 중체포미수 \\
형법 제277조 제1항, 제280조 & 중감금미수 \\
형법 제278조 & 특수체포 \\
형법 제278조 & 특수감금 \\
형법 제278조 & 특수중체포 \\
형법 제278조 & 특수중감금 \\
형법 제278조, 제280조 & 특수체포미수 \\
형법 제278조, 제280조 & 특수감금미수 \\
형법 제278조, 제280조 & 특수중체포미수 \\
형법 제278조, 제280조 & 특수중감금미수 \\
형법 제279조 & 상습체포 \\
형법 제279조 & 상습감금 \\
형법 제279조 & 상습중체포 \\
형법 제279조 & 상습중감금 \\
형법 제279조, 제280조 & 상습체포미수 \\
형법 제279조, 제280조 & 상습감금미수 \\
형법 제279조, 제280조 & 상습중체포미수 \\
형법 제279조, 제280조 & 상습중감금미수 \\
형법 제281조 제1항 & 체포치상 \\
형법 제281조 제1항 & 감금치상 \\
형법 제281조 제1항 & 중체포치상 \\
형법 제281조 제1항 & 중감금치상 \\
형법 제281조 제1항 & 체포치사 \\
형법 제281조 제1항 & 감금치사 \\
형법 제281조 제1항 & 중체포치사 \\
형법 제281조 제1항 & 중감금치사 \\
형법 제281조 제1항 & 특수존속체포 \\
형법 제281조 제1항 & 상습존속체포 \\
형법 제281조 제1항 & 특수존속체포미수 \\
형법 제281조 제1항 & 상습존속체포미수 \\
형법 제281조 제1항 & 특수존속감금 \\
형법 제281조 제1항 & 상습존속감금 \\
형법 제281조 제1항 & 특수존속감금미수 \\
형법 제281조 제1항 & 상습존속감금미수 \\
형법 제281조 제1항 & 특수중존속체포 \\
형법 제281조 제1항 & 상습중존속체포 \\
형법 제281조 제1항 & 특수중존속체포미수 \\
형법 제281조 제1항 & 상습중존속체포미수 \\
형법 제281조 제1항 & 특수중존속감금 \\
형법 제281조 제1항 & 상습중존속감금 \\
형법 제281조 제1항 & 특수중존속감금미수 \\
형법 제281조 제1항 & 상습중존속감금미수 \\
형법 제277조 제2항 & 중존속체포 \\
형법 제277조 제2항 & 중존속감금 \\
형법 제276조 제2항 & 존속체포 \\
형법 제276조 제2항 & 존속감금 \\
폭력행위등처벌에관한법률 제2조 제2항, 형법 제276조 제1항 & 폭력행위등처벌에관한법률위반(체포) \\
폭력행위등처벌에관한법률 제2조 제2항, 형법 제276조 제1항 & 폭력행위등처벌에관한법률위반(감금) \\
폭력행위등처벌에관한법률 제2조 제2항, 형법 제276조 제2항 & 폭력행위등처벌에관한법률위반(존속체포) \\
폭력행위등처벌에관한법률 제2조 제2항, 형법 제276조 제2항 & 폭력행위등처벌에관한법률위반(존속감금) \\
폭력행위등처벌에관한법률 제2조 제3항, 형법 제276조 제1항 & 폭력행위등처벌에관한법률위반(체포재범) \\
폭력행위등처벌에관한법률 제2조 제3항, 형법 제276조 제1항 & 폭력행위등처벌에관한법률위반(감금재범) \\
폭력행위등처벌에관한법률 제2조 제3항, 형법 제276조 제2항 & 폭력행위등처벌에관한법률위반(존속체포재범) \\
폭력행위등처벌에관한법률 제2조 제3항, 형법 제276조 제2항 & 폭력행위등처벌에관한법률위반(존속감금재범) \\
\hline
\end{longtable}
\renewcommand{\arraystretch}{1.3} 

\scriptsize
\begin{longtable}{p{0.47\textwidth} p{0.47\textwidth}}
\caption{List of query case types for Threat}\label{tab:korean_threat}\\
\hline
\textbf{Law} & \textbf{Crime}\\
\hline
\endfirsthead
\hline
\textbf{Law} & \textbf{Crime}\\
\hline
\endhead
\hline
\endfoot
형법 제283조 제1항 & 협박 \\
형법 제284조 & 특수협박 \\
형법 제285조 & 상습협박 \\
형법 제285조, 제284조 & 상습특수협박 \\
형법 제286조 & 협박미수 \\
형법 제284조 & 특수협박미수 \\
형법 제285조 & 상습협박미수 \\
형법 제285조, 제284조 & 상습특수협박미수 \\
형법 제283조 제2항 & 존속협박 \\
형법 제284조, 제283조 제2항 & 특수존속협박 \\
형법 제285조, 제283조 제2항 & 상습존속협박 \\
폭력행위등처벌에관한법률 제2조 제2항, 형법 제283조 제1항 & 폭력행위등처벌에관한법률위반(협박) \\
폭력행위등처벌에관한법률 제2조 제2항, 형법 제283조 제2항 & 폭력행위등처벌에관한법률위반(존속협박) \\
폭력행위등처벌에관한법률 제2조 제3항, 형법 제283조 제1항 & 폭력행위등처벌에관한법률위반(협박재범) \\
폭력행위등처벌에관한법률 제2조 제3항, 형법 제283조 제2항 & 폭력행위등처벌에관한법률위반(존속협박재범) \\
\hline
\end{longtable}

\renewcommand{\arraystretch}{1.3} 

\scriptsize
\begin{longtable}{p{0.47\textwidth} p{0.47\textwidth}}
\caption{List of query case types for Defamation and Insult}\label{tab:korean_defamation}\\
\hline
\textbf{Law} & \textbf{Crime}\\
\hline
\endfirsthead
\hline
\textbf{Law} & \textbf{Crime}\\
\hline
\endhead
\hline
\endfoot
형법 제307조 제1항 & 사실적시명예훼손 \\
형법 제307조 제2항 & 허위적시명예훼손 \\
형법 제309조, (제307조 제1항, 제307조 제2항) & 출판물에의한명예훼손 \\
형법 제309조, (제307조 제1항, 제307조 제2항) & 라디오에의한명예훼손 \\
형법 제311조 & 모욕 \\
정보통신망이용촉진및정보보호등에관한법률 제70조 제1항 & 정보통신망이용촉진및정보보호등에관한법률위반(사실적시명예훼손) \\
정보통신망이용촉진및정보보호등에관한법률 제70조 제2항 & 정보통신망이용촉진및정보보호등에관한법률위반(허위적시명예훼손) \\
정보통신망이용촉진및정보보호등에관한법률 제71조 제1항 & 정보통신망이용촉진및정보보호등에관한법률위반(정보통신망침해등) \\
정보통신망이용촉진및정보보호등에관한법률 제74조 제1항 제2호 & 정보통신망이용촉진및정보보호등에관한법률위반(음란물유포) \\
\hline
\end{longtable}
\renewcommand{\arraystretch}{1.3} 

\scriptsize
\begin{longtable}{p{0.47\textwidth} p{0.47\textwidth}}
\caption{List of query case types for Obstruction of business}\label{tab:korean_obs_business}\\
\hline
\textbf{Law} & \textbf{Crime}\\
\hline
\endfirsthead
\hline
\textbf{Law} & \textbf{Crime}\\
\hline
\endhead
\hline
\endfoot
형법 제314조 제1항 & 업무방해 \\
형법 제314조 제1항 & 공무집행방해 \\
형법 제314조 제1항 & 위계에의한공무집행방해 \\
형법 제314조 제1항 & 공용서류손상 \\
형법 제314조 제1항 & 공용서류은닉 \\
형법 제314조 제1항 & 공용서류무효 \\
형법 제314조 제1항 & 공용물건손상 \\
형법 제314조 제1항 & 공용물건은닉 \\
형법 제314조 제1항 & 공용물건무효 \\
형법 제314조 제1항 & 공용전자기록등손상 \\
형법 제314조 제1항 & 공용전자기록등은닉 \\
형법 제314조 제1항 & 공용전자기록등무효 \\
형법 제314조 제1항 & 특수공무집행방해치사 \\
형법 제314조 제1항 & 특수공무집행방해치상 \\
형법 제314조 제1항 & 특수공무집행방해 \\
형법 제314조 제1항 & 특수공용물건손상 \\

\hline
\end{longtable}
\renewcommand{\arraystretch}{1.3} 

\scriptsize
\begin{longtable}{p{0.47\textwidth} p{0.47\textwidth}}
\caption{List of query case types for Home invasion}\label{tab:korean_home_invasion}\\
\hline
\textbf{Law} & \textbf{Crime}\\
\hline
\endfirsthead
\hline
\textbf{Law} & \textbf{Crime}\\
\hline
\endhead
\hline
\endfoot
형법 제319조 제1항 & 주거침입 \\
형법 제319조 제1항 & 건조물침입 \\
형법 제319조 제1항 & 방실침입 \\
형법 제319조 제2항 & 퇴거불응 \\
형법 제320조 & 특수주거침입 \\
형법 제320조 & 특수건조물침입 \\
형법 제320조 & 특수방실침입 \\
형법 제320조 & 특수퇴거불응 \\
형법 제322조 & 주거침입미수 \\
형법 제322조 & 건조물침입미수 \\
형법 제322조 & 방실침입미수 \\
형법 제322조 & 퇴거불응미수 \\
형법 제320조, 제322조 & 특수주거침입미수 \\
형법 제320조, 제322조 & 특수건조물침입미수 \\
형법 제320조, 제322조 & 특수방실침입미수 \\
형법 제320조, 제322조 & 특수퇴거불응미수 \\
폭력행위등처벌에관한법률 제2조 제2항, 형법 제319조 제1항 & 폭력행위등처벌에관한법률위반(주거침입) \\
폭력행위등처벌에관한법률 제2조 제2항, 형법 제319조 제2항 & 폭력행위등처벌에관한법률위반(퇴거불응) \\
폭력행위등처벌에관한법률 제2조 제3항, 형법 제319조 제1항 & 폭력행위등처벌에관한법률위반(주거침입재범) \\
폭력행위등처벌에관한법률 제2조 제3항, 형법 제319조 제2항 & 폭력행위등처벌에관한법률위반(퇴거불응재범) \\
\hline
\end{longtable}
\renewcommand{\arraystretch}{1.3} 

\scriptsize
\begin{longtable}{p{0.47\textwidth} p{0.47\textwidth}}
\caption{List of query case types for Obstruction of right}\label{tab:korean_obs_right}\\
\hline
\textbf{Law} & \textbf{Crime}\\
\hline
\endfirsthead
\hline
\textbf{Law} & \textbf{Crime}\\
\hline
\endhead
\hline
\endfoot
형법 제323조 & 권리행사방해 \\
형법 제324조 제1항 & 강요 \\
형법 제324조 제2항 & 특수강요 \\
형법 제327조 & 강제집행면탈 \\
폭력행위등처벌에관한법률 제2조 제2항, 형법 제324조 제1항 & 폭력행위등처벌에관한법률위반(강요) \\
폭력행위등처벌에관한법률 제2조 제3항, 형법 제324조 제1항 & 폭력행위등처벌에관한법률위반(강요재범) \\

\hline
\end{longtable}
\renewcommand{\arraystretch}{1.3} 

\scriptsize
\begin{longtable}{p{0.47\textwidth} p{0.47\textwidth}}
\caption{List of query case types for Drug}\label{tab:korean_drug}\\
\hline
\textbf{Law} & \textbf{Crime}\\
\hline
\endfirsthead
\hline
\textbf{Law} & \textbf{Crime}\\
\hline
\endhead
\hline
\endfoot
마약류관리에관한법률 제2조 제2호 & 마약류관리에관한법률위반(마약) \\
마약류관리에관한법률 제5조의2 제5항 & 마약류관리에관한법률위반임시마약류마약 \\
마약류관리에관한법률 제2조 제3호 & 마약류관리에관한법률위반(향정) \\
마약류관리에관한법률 제5조의2 제5항 & 마약류관리에관한법률위반임시마약류향정 \\
마약류관리에관한법률 제2조 제4호 & 마약류관리에관한법률위반(대마) \\
마약류관리에관한법률 제5조의2 제5항 & 마약류관리에관한법률위반임시마약류대마 \\
\hline
\end{longtable}
\renewcommand{\arraystretch}{1.3} 

\scriptsize
\begin{longtable}{p{0.47\textwidth} p{0.47\textwidth}}
\caption{List of query case types for Gambling}\label{tab:korean_gambling}\\
\hline
\textbf{Law} & \textbf{Crime}\\
\hline
\endfirsthead
\hline
\textbf{Law} & \textbf{Crime}\\
\hline
\endhead
\hline
\endfoot
형법 제246조 제1항 & 도박 \\
형법 제246조 제2항 & 상습도박 \\
형법 제247조 & 도박장소개설 \\
형법 제247조 & 도박공간개설 \\
국민체육진흥법 제47조 제2호 & 국민체육진흥법위반(도박개장등) \\
국민체육진흥법 제48조 제3호, 제26조 제1항 & 국민체육진흥법위반(도박등) \\
국민체육진흥법 제48조 제4호, 제26조 제2항 제1호 & 국민체육진흥법위반(도박개장등) \\
게임산업진흥에관한법률 & 게임산업진흥에관한법률위반 \\
\hline
\end{longtable}
\renewcommand{\arraystretch}{1.3} 
\scriptsize
\begin{longtable}{p{0.47\textwidth} p{0.47\textwidth}}
\caption{List of query case types for Other criminal offenses}\label{tab:korean_other_crime}\\
\hline
\textbf{Law} & \textbf{Crime}\\
\hline
\endfirsthead
\hline
\textbf{Law} & \textbf{Crime}\\
\hline
\endhead
\hline
\endfoot
형법 제145조 & 도주 \\
형법 제151조 & 범인은닉 \\
형법 제151조 & 범인도피 \\
형법 제152조 제1항 & 위증 \\
형법 제156조 & 무고 \\
형법 제164조 제1항, 제165조, 제166조, 제167조 & 방화 \\
형법 제164조 제2항, 제164조 제1항 & 방화치상 \\
형법 제164조 제2항, 제164조 제1항 & 방화치사 \\
형법 제170조 & 실화 \\
형법 제171조 & 업무상실화 \\
형법 제174조 & 방화미수 \\
형법 제175조 & 방화예비 \\
형법 제185조 & 일반교통방해 \\
형법 제186조 & 기차교통방해 \\
형법 제186조 & 전차교통방해 \\
\hline
\end{longtable}
\renewcommand{\arraystretch}{1.3} 

\scriptsize
\begin{longtable}{p{0.47\textwidth} p{0.47\textwidth}}
\caption{List of query case types for Traffic offenses}\label{tab:korean_traffic}\\
\hline
\textbf{Law} & \textbf{Crime}\\
\hline
\endfirsthead
\hline
\textbf{Law} & \textbf{Crime}\\
\hline
\endhead
\hline
\endfoot
교통사고처리특례법 제3조 제1항, 형법 제268조 & 교통사고처리특례법위반(치사) \\
교통사고처리특례법 제3조 제1항, 형법 제268조 & 교통사고처리특례법위반(치상) \\
교통사고처리특례법 & 교통사고처리특례법위반 \\
특정범죄가중처벌등에관한법률 제5조의3 제1항 제1호 & 특정범죄가중처벌등에관한법률위반(도주치사) \\
특정범죄가중처벌등에관한법률 제5조의3 제1항 제2호 & 특정범죄가중처벌등에관한법률위반(도주치상) \\
특정범죄가중처벌등에관한법률 제5조의10 & 특정범죄가중처벌등에관한법률위반(운전자폭행등) \\
특정범죄가중처벌등에관한법률 제5조의11 제1항 & 특정범죄가중처벌등에관한법률위반(위험운전치사) \\
특정범죄가중처벌등에관한법률 제5조의11 제1항 & 특정범죄가중처벌등에관한법률위반(위험운전치상) \\
특정범죄가중처벌등에관한법률 제5조의13 & 특정범죄가중처벌등에관한법률위반(어린이보호구역치사) \\
특정범죄가중처벌등에관한법률 제5조의13 & 특정범죄가중처벌등에관한법률위반(어린이보호구역치상) \\
도로교통법 제43조 & 도로교통법위반(무면허운전) \\
도로교통법 제44조 제1항 & 도로교통법위반(음주운전) \\
도로교통법 제44조 제2항 & 도로교통법위반(음주측정거부) \\
도로교통법 제46조 & 도로교통법위반(공동위험행위) \\
도로교통법 제54조 제1항 & 도로교통법위반(사고후미조치) \\
\hline
\end{longtable}
\renewcommand{\arraystretch}{1.3} 

\scriptsize
\begin{longtable}{p{0.47\textwidth} p{0.47\textwidth}}
\caption{List of query case types for Child crime and School violence}\label{tab:korean_child_school}\\
\hline
\textbf{Law} & \textbf{Crime}\\
\hline
\endfirsthead
\hline
\textbf{Law} & \textbf{Crime}\\
\hline
\endhead
\hline
\endfoot
아동복지법 제71조 제1항 제2호, 제17조 각호 & 아동복지법위반(아동학대) \\
아동복지법 & 아동복지법위반(아동유기·방임) \\
아동복지법 & 아동복지법위반(상습아동학대) \\
아동복지법 & 아동복지법위반(상습아동유기·방임) \\
아동학대범죄의처벌등에관한특례법 제4조 제2항 & 아동학대범죄의처벌등에관한특례법위반(아동학대치사) \\
아동학대범죄의처벌등에관한특례법 제5조 & 아동학대범죄의처벌등에관한특례법위반(아동학대중상해) \\
아동학대범죄의처벌등에관한특례법 제7조 & 아동학대범죄의처벌등에관한특례법위반(아동복지시설종사자등의아동학대) \\
청소년보호법 & 청소년보호법위반 \\
학교폭력예방및대책에관한법률 & 학교폭력예방및대책에관한법률위반 \\
\hline
\end{longtable}

\renewcommand{\arraystretch}{1.3} 

\scriptsize
\begin{longtable}{p{0.47\textwidth} p{0.47\textwidth}}
\caption{List of query case types for Corporation}\label{tab:korean_corporation}\\
\hline
\textbf{Law} & \textbf{Crime}\\
\hline
\endfirsthead
\hline
\textbf{Law} & \textbf{Crime}\\
\hline
\endhead
\hline
\endfoot
부정경쟁방지법 제18조 제1항 & 부정경쟁방지및영업비밀보호에관한법률위반(영업비밀국외누설등) \\
부정경쟁방지법 제18조 제2항 & 부정경쟁방지및영업비밀보호에관한법률위반(영업비밀누설등) \\
자본시장법 & 자본시장과금융투자업에관한법률위반 \\
\hline
\end{longtable}
\renewcommand{\arraystretch}{1.3} 
\scriptsize
\begin{longtable}{p{0.47\textwidth} p{0.47\textwidth}}
\caption{List of query case types for Labor and Employment}\label{tab:korean_labor}\\
\hline
\textbf{Law} & \textbf{Crime}\\
\hline
\endfirsthead
\hline
\textbf{Law} & \textbf{Crime}\\
\hline
\endhead
\hline
\endfoot
근로기준법 제107조, 제7조 & 근로기준법위반(강제근로의금지) \\
근로기준법 제107조, 제8조 & 근로기준법위반(폭행의금지) \\
근로기준법 제107조, 제9조 & 근로기준법위반(중간착취의배제) \\
근로기준법 제107조, 제23조 제2항 & 근로기준법위반(해고등의제한) \\
근로기준법 제107조, 제40조 & 근로기준법위반(취업방해의금지) \\
근로기준법 제109조 제1항, 제36조 & 근로기준법위반(금품청산) \\
근로기준법 제109조 제1항, 제43조 & 근로기준법위반(임금지급) \\
근로기준법 제109조 제1항, 제46조 & 근로기준법위반(휴업수당지급) \\
근로기준법 제109조 제1항, 제51조의2 & 근로기준법위반(가산임금지급) \\
근로기준법 제109조 제1항, 제52조 제2항 제2호 & 근로기준법위반(가산임금지급) \\
근로기준법 제109조 제1항, 제56조 & 근로기준법위반(가산임금지급) \\
근로기준법 제109조 제1항, 제65조 & 근로기준법위반(사용금지) \\
근로기준법 제109조 제1항, 제72조 & 근로기준법위반(갱내근로의금지) \\
근로기준법 제109조 제1항, 제76조의3 제6항 & 근로기준법위반(직장내괴롭힘발생시조치) \\
근로기준법 제110조 제1호, 제10조 & 근로기준법위반(공민권행사보장) \\
근로기준법 제110조 제1호, 제22조 제1항 & 근로기준법위반(강제저금의금지) \\
근로기준법 제110조 제1호, 제26조 & 근로기준법위반(해고의예고) \\
근로기준법 제110조 제1호, 제50조 & 근로기준법위반(법정근로시간) \\
근로기준법 제110조 제1호, 제51조의2 제2항 & 근로기준법위반(휴식시간제공) \\
근로기준법 제110조 제1호, 제52조 제2항 제1호 & 근로기준법위반(휴식시간제공) \\
근로기준법 제110조 제1호, 제53조 제1항 & 근로기준법위반(연장근로의제한) \\
근로기준법 제110조 제1호, 제53조 제2항 & 근로기준법위반(연장근로의제한) \\
근로기준법 제110조 제1호, 제53조 제4항 본문 & 근로기준법위반(연장근로의제한) \\
근로기준법 제110조 제1호, 제53조 제7항 & 근로기준법위반(연장근로의제한) \\
근로기준법 제110조 제1호, 제54조 & 근로기준법위반(휴식시간제공) \\
근로기준법 제110조 제1호, 제55조 & 근로기준법위반(휴일보장) \\
근로기준법 제110조 제1호, 제59조 제2항 & 근로기준법위반(휴식시간제공) \\
근로기준법 제110조 제1호, 제60조 제1항 & 근로기준법위반(연차유급휴가지급) \\
근로기준법 제110조 제1호, 제60조 제2항 & 근로기준법위반(연차유급휴가지급) \\
근로기준법 제110조 제1호, 제60조 제4항 & 근로기준법위반(연차유급휴가지급) \\
근로기준법 제110조 제1호, 제60조 제5항 & 근로기준법위반(연차유급휴가지급) \\
근로기준법 제110조 제1호, 제69조 & 근로기준법위반(미성년자근로시간) \\
근로기준법 제110조 제1호, 제70조 제1항 & 근로기준법위반(야간근로휴일근로) \\
근로기준법 제110조 제1호, 제70조 제2항 & 근로기준법위반(야간근로휴일근로) \\
근로기준법 제110조 제1호, 제71조 & 근로기준법위반(시간외근로) \\
근로기준법 제110조 제1호, 제74조 & 근로기준법위반(임산부보호) \\
근로기준법 제110조 제1호, 제75조 & 근로기준법위반(육아시간제공) \\
근로기준법 제110조 제1호, 제104조 제2항 & 근로기준법위반(불리한처우금지) \\
근로기준법 제110조 제2호, 제53조 제5항 & 근로기준법위반(휴식시간제공) \\
근로기준법 제114조 제1호, 제6조 & 근로기준법위반(균등한처우) \\
근로기준법 제114조 제1호, 제17조 & 근로기준법위반(근로조건명시) \\
근로기준법 제114조 제1호, 제20조 & 근로기준법위반(위약예정금지) \\
근로기준법 제114조 제1호, 제21조 & 근로기준법위반(전차금상계금지) \\
근로기준법 제114조 제1호, 제22조 제2항 & 근로기준법위반(강제저금금지) \\
근로기준법 제114조 제1호, 제67조 제1항 & 근로기준법위반(근로계약대리금지) \\
근로기준법 제114조 제1호, 제67조 제3항 & 근로기준법위반(근로조건서면교부) \\
근로기준법 제114조 제1호, 제70조 제3항 & 근로기준법위반(야간근로휴일근로제한) \\
근로기준법 제114조 제1호, 제73조 & 근로기준법위반(생리휴가지급) \\
근로기준법 제114조 제1호, 제74조 제6항 & 근로기준법위반(임산부의보호) \\
근로기준법 제114조 제1호, 제94조 & 근로기준법위반(취업규칙작성변경) \\
근로기준법 제114조 제1호, 제95조 & 근로기준법위반(감급제재제한규정) \\
근로기준법 제116조 제1항, 제76조의2 & 근로기준법위반(직장내괴롭힘금지) \\
근로기준법 제116조 제2항 제2호, 제41조 & 근로기준법위반(근로자명부작성) \\
근로기준법 제116조 제2항 제2호, 제42조 & 근로기준법위반(계약서류보존) \\
근로기준법 제116조 제2항 제2호, 제48조 & 근로기준법위반(임금대장및임금내역서) \\
근로기준법 제116조 제2항 제2호, 제74조 제7항 & 근로기준법위반(임산부의보호) \\
근로기준법 제116조 제2항 제2호, 제74조 제9항 & 근로기준법위반(임산부의보호) \\
근로기준법 제116조 제2항 제2호, 제76조의3 제2항 & 근로기준법위반(직장내괴롭힘발생시조치) \\
근로기준법 제116조 제2항 제2호, 제76조의3 제4항 & 근로기준법위반(직장내괴롭힘발생시조치) \\
근로기준법 제116조 제2항 제2호, 제76조의3 제5항 & 근로기준법위반(직장내괴롭힘발생시조치) \\
근로기준법 제116조 제2항 제2호, 제76조의3 제7항 & 근로기준법위반(직장내괴롭힘발생시조치) \\
근로기준법 제116조 제2항 제2호, 제91조 & 근로기준법위반(서류보존) \\
근로기준법 제116조 제2항 제2호, 제93조 & 근로기준법위반(취업규칙작성신고) \\
근로기준법 제116조 제2항 제3호, 제51조의2 제5항 & 근로기준법위반(임금보전방안마련) \\
퇴직급여법 & 근로자퇴직급여보장법위반 \\
최저임금법 & 최저임금법위반 \\
노동조합법 제88조, 제41조 & 노동조합및노동관계조정법위반(쟁의행위제한과금지) \\
노동조합법 제89조, 제37조 & 노동조합및노동관계조정법위반(노동조합아닌자에의한쟁의행위금지) \\
노동조합법 제89조, 제38조 & 노동조합및노동관계조정법위반(노동조합의지도와책임) \\
노동조합법 제89조, 제42조 & 노동조합및노동관계조정법위반(폭력행위등의금지) \\
노동조합법 제89조, 제42조의2 & 노동조합및노동관계조정법위반(필수유지업무에대한쟁의행위제한) \\
노동조합법 제90조, 제44조 & 노동조합및노동관계조정법위반(쟁의행위기간중의임금지급요구금지) \\
노동조합법 제90조, 제77조 & 노동조합및노동관계조정법위반(긴급조정시쟁의행위중지) \\
노동조합법 제90조, 제81조 & 노동조합및노동관계조정법위반(부당노동행위) \\
노동조합법 제91조, 제38조 & 노동조합및노동관계조정법위반(노동조합의지도와책임) \\
노동조합법 제91조, 제41조 & 노동조합및노동관계조정법위반(쟁의행위제한금지) \\
노동조합법 제91조, 제42조 & 노동조합및노동관계조정법위반(폭력행위등의금지) \\
노동조합법 제91조, 제43조 & 노동조합및노동관계조정법위반(사용자의채용제한) \\
노동조합법 제91조, 제46조 & 노동조합및노동관계조정법위반(직장폐쇄) \\
노동조합법 제91조, 제63조 & 노동조합및노동관계조정법위반(중재시쟁의행위금지) \\
노동조합법 제92조, 제31조 & 노동조합및노동관계조정법위반(단체협약준수) \\
노동조합법 제93조, 제7조 & 노동조합및노동관계조정법위반(노동조합의보호요건) \\
노동조합법 제93조, 제21조, 제31조 & 노동조합및노동관계조정법위반(시정명령) \\
노동조합법 제96조, 제14조 & 노동조합및노동관계조정법위반(서류비치) \\
노동조합법 제96조, 제27조 & 노동조합및노동관계조정법위반(자료제출) \\
노동조합법 제96조, 제46조 & 노동조합및노동관계조정법위반(직장폐쇄신고) \\
노동조합법 제96조, 제13조, 제28조제2항, 제31조 제2항 & 노동조합및노동관계조정법위반(신고의무) \\

\hline
\end{longtable}
\renewcommand{\arraystretch}{1.3} 

\scriptsize
\begin{longtable}{p{0.47\textwidth} p{0.47\textwidth}}
\caption{List of query case types for Industrial accident and Serious accident}\label{tab:korean_industry_serious}\\
\hline
\textbf{Law} & \textbf{Crime}\\
\hline
\endfirsthead
\hline
\textbf{Law} & \textbf{Crime}\\
\hline
\endhead
\hline
\endfoot
산업안전법 제167조, 제38조 제1,2,3항, 제166조의2 & 산업안전보건법위반(안전조치) \\
산업안전법 제167조, 제39조 제1항 & 산업안전보건법위반(보건조치) \\
산업안전법 제167조, 제63조 & 산업안전보건법위반(안전보건조치) \\
산업안전법 제168조, 제38조 제1,2,3항 & 산업안전보건법위반(안전조치) \\
산업안전법 제168조, 제39조 제1항 & 산업안전보건법위반(보건조치) \\
산업안전법 제168조, 제51조 & 산업안전보건법위반(작업중지의무) \\
산업안전법 제168조, 제54조 제1항 & 산업안전보건법위반(사업주의조치) \\
산업안전법 제168조, 제117조 제1항 & 산업안전보건법위반(유해위험물질제조등금지) \\
산업안전법 제168조, 제118조 제1항 & 산업안전보건법위반(무허가유해위험물질제조) \\
산업안전법 제168조, 제122조 제1항 & 산업안전보건법위반(석면해체제거) \\
산업안전법 제168조, 제157조 제3항 & 산업안전보건법위반(불리한처우금지) \\
산업안전법 제169조, 제45조 제1항, 제46조 제5항, 제53조 제1항, 제87조 제2항, 제118조 제4항, 제119조 제4항, 제131조 제1항 & 산업안전보건법위반(명령위반) \\
산업안전법 제170조, 제41조제3항 & 산업안전보건법위반(불리한처우금지) \\
산업안전법 제170조, 제56조제3항 & 산업안전보건법위반(중대재해발생현장훼손) \\
산업안전법 제170조, 제57조제1항 & 산업안전보건법위반(산업재해사실은폐교사공모) \\
산업안전법 제170조, 제153조 & 산업안전보건법위반(자격증등록증대여) \\
산업안전법 제171조, 제125조제6항, 제132조제4항 & 산업안전보건법위반(조치의무) \\
산업안전법 제172조, 제64조 제1항 제7호, 제64조 제1항 제8호, 제64조 제2항 & 산업안전보건법위반(산업재해예방조치의무) \\
산재보험법 & 산업재해보상보험법위반 \\
중대재해처벌법 & 중대재해처벌등에관한법률위반 \\
\hline
\end{longtable}
\renewcommand{\arraystretch}{1.3} 

\scriptsize
\begin{longtable}{p{0.47\textwidth} p{0.47\textwidth}}
\caption{List of query case types for Medical or Food drug}\label{tab:korean_medical_food_drug}\\
\hline
\textbf{Law} & \textbf{Crime}\\
\hline
\endfirsthead
\hline
\textbf{Law} & \textbf{Crime}\\
\hline
\endhead
\hline
\endfoot
약사법 제93조 제1항 제1호, 제6조 제3항 & 약사법위반(약사면허대여) \\
약사법 제93조 제1항 제2호, 제20조 제1항 & 약사법위반(약국개설금지) \\
약사법 제93조 제1항 제3호, 제23조 제1항 & 약사법위반(무면허의약품조제) \\
약사법 제93조 제1항 제4호, 제31조 제1항, 제2항, 제3항, 제4항, 제9항 & 약사법위반(허가,신고미이행) \\
약사법 제93조 제1항 제5호, 제42조 제1항 & 약사법위반(무허가,무신고의약품수입등) \\
약사법 제93조 제1항 제7호, 제44조 제1항 & 약사법위반((약국개설자가아님에도)의약품판매등) \\
약사법 제93조 제1항 제10호, 제61조 제1항 제1호 & 약사법위반(위조의약품판매등) \\
약사법 제93조 제1항 제10호, 제61조 제1항 제2호 & 약사법위반(무허가,무신고의약품판매등) \\
약사법 제93조 제1항 제10호, 제61조 제2항 & 약사법위반(의약품유사광고등) \\
약사법 제93조 제1항 제10호, 제66조~ & 약사법위반(의약외품) \\
약사법 제94조 제1항 제2호, 제24조 제2항 & 약사법위반(담합행위) \\
약사법 제94조 제1항 제5호의4, 제47조 제2항 & 약사법위반(경제적이익등제공) \\
약사법 제94조 제1항 제5호의4, 제47조 제6항 & 약사법위반(경제적이익등취득) \\
약사법 제94조 제1항 제8호, 제50조 제1항 & 약사법위반(점포이외의장소에서의약품판매) \\
약사법 제94조 제1항 제9호, 제62조 & 약사법위반(유해/위해의약품판매등) \\
약사법 제94조 제1항 제9호, 제66조, 제62조 & 약사법위반(유해/위해의약외품판매등) \\
약사법 제95조 제1항 제2호, 제21조 제1항 & 약사법위반(약국이중개설) \\
약사법 제95조 제1항 제8호, 제47조 제1항, 제4항, 제7항 & 약사법위반(유통체계확립및판매질서유지의무위반) \\
약사법 제95조 제1항 제10호, 제60조 & 약사법위반(기재금지사항) \\
약사법 제95조 제1항 제10호, 제68조 & 약사법위반(의약품과장광고등) \\
식품위생법 제94조 제1항 제1호, 제4조 & 식품위생법위반(위해식품등의판매등) \\
식품위생법 제94조 제1항 제3호, 제37조 제1항 & 식품위생법위반(무허가유흥주점) \\
식품위생법 제95조 제1호, 제7조 제4항 & 식품위생법위반(식품또는식품첨가물에관한기준및규격) \\
식품위생법 제95조 제1호, 제9조 제4항 & 식품위생법위반(기구및용기포장에관한기준및규격) \\
식품위생법 제95조 제2호의2, 제37조 제5항 & 식품위생법위반(영업허가등(허가,등록,신고)) \\
식품위생법 제96조, 제51조, 제52조 & 식품위생법위반(조리사미배치,영양사미채용) \\
식품위생법 제97조 제1호, 제37조 제4항 & 식품위생법위반(무허가,무신고영업등) \\
식품위생법 제97조 제4호, 제36조 & 식품위생법위반(시설기준위반) \\
식품위생법 제97조 제6호, 제44조 제1항 & 식품위생법위반(영업자등준수사항위반) \\
식품위생법 제97조 제7호, 제75조 제1항 & 식품위생법위반(영업정지명령위반) \\
식품위생법 제98조 제1호, 제44조 제3항 & 식품위생법위반(접객행위알선) \\

\hline
\end{longtable}
\renewcommand{\arraystretch}{1.3} 

\scriptsize
\begin{longtable}{p{0.47\textwidth} p{0.47\textwidth}}
\caption{List of query case types for Fair trade}\label{tab:korean_fair_trade}\\
\hline
\textbf{Law} & \textbf{Crime}\\
\hline
\endfirsthead
\hline
\textbf{Law} & \textbf{Crime}\\
\hline
\endhead
\hline
\endfoot
공정거래법 제6조, 제5조 제1항 제1호, 제124조 제1항 제1호 & 독점규제및공정거래에관한법률위반(시장지배적지위남용행위-가격남용행위) \\
공정거래법 제6조, 제5조 제1항 제2호, 제124조 제1항 제1호 & 독점규제및공정거래에관한법률위반(시장지배적지위남용행위-출고조절행위) \\
공정거래법 제6조, 제5조 제1항 제3호, 제124조 제1항 제1호 & 독점규제및공정거래에관한법률위반(시장지배적지위남용행위-사업활동방해행위) \\
공정거래법 제6조, 제5조 제1항 제4호, 제124조 제1항 제1호 & 독점규제및공정거래에관한법률위반(시장지배적지위남용행위-진입제한행위) \\
공정거래법 제6조, 제5조 제1항 제5호, 제124조 제1항 제1호 & 독점규제및공정거래에관한법률위반(시장지배적지위남용행위-경쟁사업자배제또는소비자이익의저해행위) \\
공정거래법 제13조, 제9조, 제124조 제1항 제2호 & 독점규제및공정거래에관한법률위반(경쟁제한적기업결합행위) \\
공정거래법 제19조, 제124조 제1항 제5호 & 독점규제및공정거래에관한법률위반(상호출자제한기업집단의지주회사설립제한위반행위) \\
공정거래법 제20조, 제124조 제1항 제6호 & 독점규제및공정거래에관한법률위반(일반지주회사의금융회사주식소유제한위반행위) \\
공정거래법 제21조, 제124조 제1항 제7호 & 독점규제및공정거래에관한법률위반(상호출자금지위반행위) \\
공정거래법 제24조, 제124조 제1항 제8호 & 독점규제및공정거래에관한법률위반(채무보증제한위반행위) \\
공정거래법 제25조, 제124조 제1항 제3호 & 독점규제및공정거래에관한법률위반(금융보험사의결권제한위반행위) \\
공정거래법 제17조, 제18조, 제124조 제1항 제4호, 제126조 제1호, 제126조 제2호 & 독점규제및공정거래에관한법률위반(지주회사등의행위제한위반행위) \\
공정거래법 제26조, 제130조 제1항 제4호, 제37조 제1항 제7호 & 독점규제및공정거래에관한법률위반(대규모내부거래에대한이사회의결및공시제도위반) \\
공정거래법 제27조, 제130조 제1항 제4호, 제37조 제1항 제7호 & 독점규제및공정거래에관한법률위반(비상장회사등의중요사항공시제도위반행위) \\
공정거래법 제28조, 제130조 제1항 제4호, 제37조 제1항 제7호 & 독점규제및공정거래에관한법률위반(기업집단현황공시제도위반행위) \\
공정거래법 제22조, 제124조 제1항 제7호 & 독점규제및공정거래에관한법률위반(순환출자금지위반행위) \\
공정거래법 제23조, 제124조 제1항 제3호 & 독점규제및공정거래에관한법률위반(순환출자에대한의결권제한위반행위) \\
공정거래법 제40조 제1항 제1호, 제124조 제1항 제9호 & 독점규제및공정거래에관한법률위반(부당공동행위-가격의결정,유지,변경행위) \\
공정거래법 제40조 제1항 제2호, 제124조 제1항 제9호 & 독점규제및공정거래에관한법률위반(부당공동행위-거래조건및대금지급조건설정행위) \\
공정거래법 제40조 제1항 제3호, 제124조 제1항 제9호 & 독점규제및공정거래에관한법률위반(부당공동행위-거래제한행위) \\
공정거래법 제40조 제1항 제4호, 제124조 제1항 제9호 & 독점규제및공정거래에관한법률위반(부당공동행위-시장분할행위) \\
공정거래법 제40조 제1항 제5호, 제124조 제1항 제9호 & 독점규제및공정거래에관한법률위반(부당공동행위-설비제한행위) \\
공정거래법 제40조 제1항 제6호, 제124조 제1항 제9호 & 독점규제및공정거래에관한법률위반(부당공동행위-상품의종류,규격제한행위) \\
공정거래법 제40조 제1항 제7호, 제124조 제1항 제9호 & 독점규제및공정거래에관한법률위반(부당공동행위-영업의주요부문공동관리행위) \\
공정거래법 제40조 제1항 제8호, 제124조 제1항 제9호 & 독점규제및공정거래에관한법률위반(부당공동행위-입찰담합행위) \\
공정거래법 제40조 제1항 제9호, 제124조 제1항 제9호 & 독점규제및공정거래에관한법률위반(부당공동행위-정보교환공동행위) \\
공정거래법 제51조 제1항 제1호, 제124조 제1항 제12호, 제53조, 제52조 & 독점규제및공정거래에관한법률위반(사업자단체금지행위-부당한공동행위) \\
공정거래법 제51조 제1항 제2호, 제53조, 제52조 & 독점규제및공정거래에관한법률위반(사업자단체금지행위-사업자수제한행위) \\
공정거래법 제51조 제1항 제3호, 제125조 제5호, 제53조, 제52조 & 독점규제및공정거래에관한법률위반(사업자단체금지행위-사업활동방해행위) \\
공정거래법 제51조 제1항 제4호, 제53조, 제52조 & 독점규제및공정거래에관한법률위반(사업자단체금지행위-불공정거래행위) \\
공정거래법 제45조 제1항 제1호 & 독점규제및공정거래에관한법률위반(일반불공정거래행위-거래거절행위) \\
공정거래법 제45조 제1항 제2호 & 독점규제및공정거래에관한법률위반(일반불공정거래행위-차별적취급행위) \\
공정거래법 제45조 제1항 제3호 & 독점규제및공정거래에관한법률위반(일반불공정거래행위-경쟁사업자배제행위) \\
공정거래법 제45조 제1항 제4호, 제125조 제4호 & 독점규제및공정거래에관한법률위반(일반불공정거래행위-부당한고객유인행위) \\
공정거래법 제45조 제1항 제5호, 제125조 제4호 & 독점규제및공정거래에관한법률위반(일반불공정거래행위-거래강제행위) \\
공정거래법 제45조 제1항 제6호, 제125조 제4호 & 독점규제및공정거래에관한법률위반(일반불공정거래행위-거래상지위남용행위) \\
공정거래법 제45조 제1항 제7호 & 독점규제및공정거래에관한법률위반(일반불공정거래행위-구속조건부거래행위) \\
공정거래법 제45조 제1항 제8호, 제125조 제4호 & 독점규제및공정거래에관한법률위반(일반불공정거래행위-사업활동방해행위) \\
공정거래법 제45조 제1항 제9호, 령 별표2 제9호 가목, 제124조 제1항 제10호 & 독점규제및공정거래에관한법률위반(부당지원행위-부당한자금지원행위) \\
공정거래법 제45조 제1항 제9호, 령 별표2 제9호 나목, 제124조 제1항 제10호 & 독점규제및공정거래에관한법률위반(부당지원행위-부당한자산상품등지원행위) \\
공정거래법 제45조 제1항 제9호, 령 별표2 제9호 다목, 제124조 제1항 제10호 & 독점규제및공정거래에관한법률위반(부당지원행위-부당한인력지원행위) \\
공정거래법 제45조 제1항 제9호, 령 별표2 제9호 라목, 제124조 제1항 제10호 & 독점규제및공정거래에관한법률위반(부당지원행위-부당한거래단계추가행위) \\
공정거래법 제48조, 제124조 제1항 제11호 & 독점규제및공정거래에관한법률위반(보복조치행위) \\
공정거래법 제47조, 제124조 제1항 제10호 & 독점규제및공정거래에관한법률위반(특수관계인에대한부당한이익제공행위) \\
공정거래법 제81조 제2항, 제124조 제1항 제13호 & 독점규제및공정거래에관한법률위반(조사시폭언폭행등행위) \\
공정거래법 제30조, 제126조 제3호 & 독점규제및공정거래에관한법률위반(주식소유현황신고의무위반행위) \\
표시광고법 & 표시·광고의공정화에관한법률위반 \\

\hline
\end{longtable}
\renewcommand{\arraystretch}{1.3} 

\scriptsize
\begin{longtable}{p{0.47\textwidth} p{0.47\textwidth}}
\caption{List of query case types for Finance or Insurance}\label{tab:korean_finance_insurance}\\
\hline
\textbf{Law} & \textbf{Crime}\\
\hline
\endfirsthead
\hline
\textbf{Law} & \textbf{Crime}\\
\hline
\endhead
\hline
\endfoot
특정금융거래정보의보고및이용등에관한법률 & 특정금융거래정보의보고및이용등에관한법률위반 \\
금융실명거래및비밀보장에관한법률 & 금융실명거래및비밀보장에관한법률위반 \\
대부업법 제3조, 제3조의2, 제9조의2, 제19조 제1항 & 대부업등의등록및금융이용자보호에관한법률(미등록대부업및광고행위) \\
대부업법 제5조의2 제5항, 제19조 제2항 제1호 & 대부업등의등록및금융이용자보호에관한법률(타인명의대부행위) \\
대부업법 제8조, 제11조 제1항, 제19조 제2항 제2호 & 대부업등의등록및금융이용자보호에관한법률(이자율제한행위) \\
대부업법 제9조의3 제2항, 제19조 제2항 제3호 & 대부업등의등록및금융이용자보호에관한법률(허위과장광고금지행위) \\
대부업법 제9조의9 제1항, 제2항, 제19조 제2항 제4호 & 대부업등의등록및금융이용자보호에관한법률(대부업이용자의개인정보보호처리행위) \\
대부업법 제9조의9 제3항, 제19조 제2항 제5호 & 대부업등의등록및금융이용자보호에관한법률(대부업이용자의개인정보제공보관등행위) \\
대부업법 제10조 제1항, 제7항, 제19조 제3항 & 대부업등의등록및금융이용자보호에관한법률(신용공여행위) \\
대부업법 제5조의2 제4항, 제19조 제4항 제1호 & 대부업등의등록및금융이용자보호에관한법률(상호사용행위) \\
대부업법 제7조 제3항, 제19조 제4항 제2호 & 대부업등의등록및금융이용자보호에관한법률(서류목적외사용행위) \\
대부업법 제9조의4 제1항, 제2항, 제19조 제4항 제4호 & 대부업등의등록및금융이용자보호에관한법률(불법사금융업자로부터의채권양수및추심,대부중개행위) \\
대부업법 제9조의4 제3항, 제19조 제4항 제5호 & 대부업등의등록및금융이용자보호에관한법률(불법사금융업자에의채권양도행위) \\
대부업법 제11조의2 제1항, 제2항, 제19조 제4항 제6호 & 대부업등의등록및금융이용자보호에관한법률(대부중개행위) \\
대부업법 제11조의2 제3항, 제19조 제4항 제7호 & 대부업등의등록및금융이용자보호에관한법률(중개수수료초과지급행위) \\
대부업법 제11조의2 제6항, 제19조 제4항 제9호 & 대부업등의등록및금융이용자보호에관한법률(중개수수료수취행위) \\
보험사기방지특별법 & 보험사기방지특별법위반 \\
유사수신행위의규제에관한법률 & 유사수신행위의규제에관한법률위반 \\
전자금융거래법 & 전자금융거래법위반 \\
범죄수익은닉의규제및처벌등에관한법률 & 범죄수익은닉의규제및처벌등에관한법률위반 \\

\hline
\end{longtable}
\renewcommand{\arraystretch}{1.3} 

\scriptsize
\begin{longtable}{p{0.47\textwidth} p{0.47\textwidth}}
\caption{List of query case types for Tax or Administrative or Constitutional Law}\label{tab:korean_tax}\\
\hline
\textbf{Law} & \textbf{Crime}\\
\hline
\endfirsthead
\hline
\textbf{Law} & \textbf{Crime}\\
\hline
\endhead
\hline
\endfoot
변호사법 & 변호사법위반 \\
조세범처벌법 & 조세범처벌법위반 \\
국가보안법 & 국가보안법위반 \\
출입국관리법 & 출입국관리법위반 \\
주민등록법 & 주민등록법위반 \\
전기통신사업법 & 전기통신사업법위반 \\
공직선거법 & 공직선거법위반 \\
국민체육진흥법 & 국민체육진흥법위반 \\
도시및주거환경정비법 & 도시및주거환경정비법위반 \\
집회및시위에관한법률 & 집회및시위에관한법률위반 \\
공인중개사법 & 공인중개사법위반 \\
건설산업기본법 & 건설산업기본법위반 \\
보조금관리에관한법률 & 보조금관리에관한법률위반 \\
성폭력범죄의처벌및피해자보호등에관한법률 & 성폭력범죄의처벌및피해자보호등에관한법률위반 \\

\hline
\end{longtable}
\renewcommand{\arraystretch}{1.3} 

\scriptsize
\begin{longtable}{p{0.47\textwidth} p{0.47\textwidth}}
\caption{List of query case types for Sexual morality}\label{tab:korean_sexual_morality}\\
\hline
\textbf{Law} & \textbf{Crime}\\
\hline
\endfirsthead
\hline
\textbf{Law} & \textbf{Crime}\\
\hline
\endhead
\hline
\endfoot
형법 제245조 & 공연음란 \\
\hline
\end{longtable}

\renewcommand{\arraystretch}{1.3} 

\scriptsize
\begin{longtable}{p{0.47\textwidth} p{0.47\textwidth}}
\caption{List of query case types for Misdemeanor}\label{tab:korean_misdemeanor}\\
\hline
\textbf{Law} & \textbf{Crime}\\
\hline
\endfirsthead
\hline
\textbf{Law} & \textbf{Crime}\\
\hline
\endhead
\hline
\endfoot
경범죄처벌법위반 & 경범죄처벌법위반 \\
\hline
\end{longtable}
\renewcommand{\arraystretch}{1.3} 

\scriptsize
\begin{longtable}{p{0.47\textwidth} p{0.47\textwidth}}
\caption{List of query case types for IT or Privacy offenses}\label{tab:korean_it_privacy}\\
\hline
\textbf{Law} & \textbf{Crime}\\
\hline
\endfirsthead
\hline
\textbf{Law} & \textbf{Crime}\\
\hline
\endhead
\hline
\endfoot
개인정보보호법 & 개인정보보호법위반 \\ 
통신비밀보호법 & 통신비밀보호법위반 \\
\hline
\end{longtable}

\renewcommand{\arraystretch}{1.3} 

\scriptsize
\begin{longtable}{p{0.47\textwidth} p{0.47\textwidth}}
\caption{List of query case types for Bribery}\label{tab:korean_bribery}\\
\hline
\textbf{Law} & \textbf{Crime}\\
\hline
\endfirsthead
\hline
\textbf{Law} & \textbf{Crime}\\
\hline
\endhead
\hline
\endfoot
형법 제129조 제1항 & 뇌물수수 \\
형법 제133조 제1항 & 뇌물공여 \\
특정범죄가중처벌등에관한법률 제2조, (형법 제129조, 제130조, 제132조) & 특정범죄가중처벌등에관한법률위반(뇌물) \\
\hline
\end{longtable}

\renewcommand{\arraystretch}{1.3} 

\scriptsize
\begin{longtable}{p{0.47\textwidth} p{0.47\textwidth}}
\caption{List of query case types for Intellectual property rights}\label{tab:korean_intellecual}\\
\hline
\textbf{Law} & \textbf{Crime}\\
\hline
\endfirsthead
\hline
\textbf{Law} & \textbf{Crime}\\
\hline
\endhead
\hline
\endfoot
상표법 & 상표법위반 \\ 
저작권법 & 저작권법위반 \\
특허법 & 특허법위반 \\
\hline
\end{longtable}
\renewcommand{\arraystretch}{1.3} 

\scriptsize
\begin{longtable}{p{0.47\textwidth} p{0.47\textwidth}}
\caption{List of query case types for Military duty law}\label{tab:korean_military_duty}\\
\hline
\textbf{Law} & \textbf{Crime}\\
\hline
\endfirsthead
\hline
\textbf{Law} & \textbf{Crime}\\
\hline
\endhead
\hline
\endfoot
병역법 & 병역법위반 \\
군형법 & 군형법위반 \\
\hline
\end{longtable}
\renewcommand{\arraystretch}{1.3} 

\scriptsize
\begin{longtable}{p{0.47\textwidth} p{0.47\textwidth}}
\caption{List of query case types for Car-related law}\label{tab:korean_car}\\
\hline
\textbf{Law} & \textbf{Crime}\\
\hline
\endfirsthead
\hline
\textbf{Law} & \textbf{Crime}\\
\hline
\endhead
\hline
\endfoot
자동차손해배상보장법 & 자동차손해배상보장법위반 \\
자동차관리법 & 자동차관리법위반 \\
\hline
\end{longtable}


\end{document}